\documentclass[11pt]{amsart}
\usepackage{amsmath,amssymb,epsfig}
\usepackage{framed}
\newcommand{\bfi}{\bfseries\itshape}
\newcommand{\rem}[1]{}

\DeclareMathOperator{\ad}{ad}

\newcommand{\lform}[2]{{\big( {#1} \big|\, {#2}\big)}}
\newcommand{\Lform}[2]{{\Big( {#1} \Big|\, {#2}\Big)}}
\newcommand{\scp}[2]{{\big\langle {#1}\, , \, {#2}\big\rangle}}
\newcommand{\prt}{\partial}
\newcommand{\sig}{\sigma}
\newcommand{\mR}{{\mathbb{R}}}
\newcommand{\id}{{\mathrm{id}}}
\newcommand{\ti}{\times}
\newcommand{\Ad}{\text{Ad}}
\renewcommand{\ad}{\text{ad}}
\newcommand{\de}{\delta}
\newcommand{\om}{\omega}
\newcommand{\al}{\alpha}
\newcommand{\ga}{\gamma}
\providecommand{\th}{}
\renewcommand{\th}{\theta}
\newcommand{\la}{\lambda}
\newcommand{\be}{\beta}
\newcommand{\Om}{\Omega}
\newcommand{\mg}{{\mathfrak g}}
\newcommand{\CX}{{\mathcal X}}
\newcommand{\cst}{\text{cst}}
\newcommand{\ze}{\zeta}
\newtheorem{theorem}{Theorem}
\newtheorem{definition}{Definition}

\title{
The Euler-Poincar\'e theory 
of Metamorphosis
}
\author{
Darryl D Holm}
\address{Department of Mathematics
\\Imperial College London SW7 2AZ, UK
\\ and
Computer and Computational Science
\\Los Alamos National Laboratory,
\\ MS D413 Los Alamos, NM 87545, USA
}
\email{d.holm@ic.ac.uk, dholm@lanl.gov}

\author{Alain Trouv{\'e}}
\address[Alain Trouv{\'e}]{CMLA ENS Cachan, CNRS, UniverSud,\\ 61, avenue du Pr\'esident Wilson\\ F-94 230
Cachan CEDEX} 
\email{trouve@cmla.ens-cachan.fr}

\author{Laurent Younes}
\address[Laurent Younes]{Center for Imaging Science
\\The Johns Hopkins University
\\3400 N-Charles Street 
\\Baltimore, MD21218-2686, USA 
}
\email{laurent.younes@jhu.edu}

\date{\today}
\begin{document}


\begin{abstract}
In the pattern matching approach to imaging science, the process of
``metamorphosis'' is template matching with dynamical templates \cite{ty05b}. Here, we
recast the metamorphosis equations of \cite{ty05b} into the Euler-Poincar\'e variational
framework of \cite{HoMaRa1998} and show that the metamorphosis equations contain the
equations for a perfect complex fluid \cite{Ho2002}. This result connects the ideas
underlying the process of metamorphosis in image matching to the physical concept of order
parameter in the theory of complex fluids. After developing the
general theory, we reinterpret various examples, including point set,
image and density metamorphosis. We finally discuss the issue of
matching measures with metamorphosis, for which we provide existence
theorems for the initial and boundary value problems.
\end{abstract}

\thanks{
The work of D. D. Holm was partially supported  by the US Department
of Energy, Office of Science, Applied Mathematical Research, and the
Royal Society of London Wolfson Research Merit Award.\\ 
The work of Laurent Younes was partially supported by NSF DMS-0456253. \\
D. D. Holm is grateful for stimulating discussions with C. Tronci.}

\maketitle

\section{Overview}

Pattern matching is an important component of imaging science, with
privileged applications in computerized anatomical analysis of medical
images (computational anatomy) \cite{bb82,boo89,gram99,tog99}. When
comparing images, the purpose is to find, based on the conservation
of photometric cues, an optimal nonrigid alignment between the
images. In this
context, diffeomorphic pattern matching methods have been developed,
based on this principle, and on the additional goal of defining a
(Riemannian) metric structure on spaces of deformable objects
\cite{dgm98,tro98}. They
have found multiple applications in medical imaging
\cite{jm00,bmty05,gvm03,vmty04,gty04,vg05}, where the objects of
interest include images, landmarks, measures (modeling unlabeled
point sets) and currents (modeling curves and surfaces). These methods
address the registration problem by solving a variational problem of
the form
\begin{equation}
\label{eq:lddmm}
\text{Find } \mathop{argmin}\Big(d(\id,g)^2 + \mathrm{Error term}(g.n_\mathit{temp},
n_\mathit{targ})\Big)
\end{equation}
over all diffeomorphisms $g$, where $n_\mathit{temp}$ and
$n_\mathit{targ}$ are the compared objects (usually referred to as the
template and the target), $(g,n) \mapsto g.n$ is the action of
diffeomorphisms on the objects and $d$ is a right-invariant Riemannian
distance on diffeomorphisms. This problem therefore directly relates
to geodesic in groups of diffeomorphisms, namely to the EPDiff
equation \cite{arn66,mr99,HoMaRa1998}, and the conserved initial
momentum that specifies the solution has been used in statistical
studies in order to provide anatomical characterizations of mental
disorders \cite{qym07,wbrc07}.

One of the issues in problems formulated as \eqref{eq:lddmm} is that
the error term breaks the metric aspects inherited from the distance $d$
on diffeomorphisms. This model has an inherent template vs. target
asymmetry, which is not always justified. With the purpose of
designing a fully metric approach to the template matching problem,
{\em Metamorphoses} have been introduced in \cite{my01}, and
formalized and studied in \cite{ty05b,ty05}. They provide interesting
pattern matching alternatives to \eqref{eq:lddmm}, in a completely
metric framework. In this paper, we pursue the following twofold goal:
(i) Provide a generic Lagrangian formulation for metamorphoses, that
includes the Riemannian formalism introduced in \cite{ty05b}, and (ii)
Study a new form of metamorphoses, adapted to the deformation of
measures.

We start with (i), for which our point of view will be fairly
abstract. We consider a manifold, $N$ which is acted upon by a Lie
group $G$: $N$ contains what we can refer to as ``deformable
objects'' and $G$ is the group of deformations, which is the group of
diffeomorphisms in our applications. We will review several examples
for the space $N$ later on in this paper.

\begin{definition}
A {\bfi
metamorphosis} \cite{ty05b} is a pair of curves $(g_t,\,\eta_t)\in G \times N$ parameterized by time $t$,
with $g_0 = \id$. Its {\bfi image} is the curve $n_t\in N$ defined by the action 
$n_t = g_t.\eta_t$.
The quantities $g_t$ and $\eta_t$ are called the {\bfi deformation part} of the
metamorphosis, and its {\bfi template part}, respectively. When $\eta_t$ is
constant, the metamorphosis is said to be a {\bfi pure
deformation}. In the general case, the image is a combination of a
deformation and template variation.
\end{definition}

In \cite{ty05b}, metamorphoses were used to modify an original
Riemannian metric on $N$ by including a deformation component in the
geodesic evolution. In this paper, we generalize the approach to a
generic Lagrangian formulation, and apply the 
Euler-Poincar\'e variational framework
\cite{HoMaRa1998} to derive evolution equations. More specific statements
on these equations (for example, regarding the existence and uniqueness of
solutions) require additional assumptions on $G$ and the space $N$ of
deformed objects. In the second part of this paper, we will review the
case in which $N$ is a space of linear forms on a Hilbert space of smooth
functions, which will allow us to define metamorphoses between
measures. 

The next section provides notation and definitions related to the
general problem of metamorphoses.


\section{Notation and Lagrangian formulation}
We will use either letters $\eta$ or $n$ to denote elements of $N$,
the former being associated to the template part of a metamorphosis,
and the latter to its image. 

The variational problem we shall study optimizes over metamorphoses
$(g_t, \eta_t)$ by minimizing, for some Lagrangian $L$,
\begin{equation}
\label{eq:lag}
\int_0^1 L(g_t, \dot g_t, \eta_t, \dot\eta_t) dt
\end{equation}
with fixed boundary conditions for the initial and final images $n_0$
and $n_1$ (with $n_t = g_t\eta_t$) and $g_0 = \id_G$ (so only the
images are constrained at the end-points, with the additional
normalization $g_0 = \id$).

Let $\mathfrak
g$ denote the Lie algebra of $G$. We will consider  Lagrangians
defined on $TG\times TN$, that satisfy the following invariance 
conditions: there exists a function $\ell$ defined on $\mathfrak g \ti
TN$ such that
$$
L(g, U_g, \eta, \xi_\eta) = \ell(U_gg^{-1}, g\eta, g\xi_\eta).
$$
In other terms, $L$ is invariant by the right action of $G$ on $G\times
N$ defined by $(g, \eta)h = (gh, h^{-1}\eta)$. 

For a metamorphosis $(g_t, \eta_t)$, we therefore have, letting $u_t = \dot g_t
g_t^{-1}$, $n_t = g_t\eta_t$ and $\nu_t = g_t \dot \eta_t$,
$$
L(g_t, \dot g_t, \eta_t, \dot\eta_t) = \ell(u_t, n_t, \nu_t).
$$

\bigskip

The Lie derivative with respect to a vector field $X$ will be denoted
$\mathcal L_X$. The Lie algebra of $G$ is identified with the set of
right invariant vector fields $U_g = ug$, $u\in T_\id G= \mathfrak g$, $g\in G$,
and we will use the notation $\mathcal L_u = \mathcal L_U$. 

The Lie
bracket $[u,v]$ on $\mathfrak g$ is defined by
$$
\mathcal L_{[u, v]} = -(\mathcal L_u\mathcal L_v - \mathcal
L_v\mathcal L_u)
$$
and the associated adjoint operator is $\ad_u v = [u,v]$.
Letting $I_g(h) = ghg^{-1}$ and $Ad_vg = \mathcal L_v I_g(\id)$, we
also have $\ad_u v = \mathcal L_u(\Ad_v)(\id)$. When $G$ is a group of
diffeomorphisms, this yields $ad_u v = du\, v -dv\, u$. 

The pairing between a linear form $l$ and a vector $u$ will be denoted
$\lform{l}{u}$. Duality with respect to this pairing will be denoted
with a $*$ exponent. 

When $G$ acts on a manifold $\tilde N$, the $\diamond$ operator is defined on
$T\tilde N^* \times
\tilde N$ and takes values in $\mathfrak g^*$. It is defined by 
$$
\lform{\delta \diamond \tilde n}{u} = - \lform{\delta}{u\tilde n}.
$$

\section{Euler Equations}

We compute the Euler equations associated with the minimization of
$$
\int_0^1 \ell(u_t, n_t, \nu_t) dt
$$
with fixed boundary conditions $n_0$ and $n_1$. We therefore consider
variations $\de u$ and $\om = \de n$. The variation $\de \nu$ can be
obtained from $n = g\eta$ and $\nu = g\dot \eta$ yielding $\dot n =
\nu + un$ and $\dot\om = \delta \nu + u\om + \delta u n$. Here and in
the following of this paper, we assume that computations are performed
in a local chart on $TN$ with respect to which we take partial derivatives.

We
therefore have
$$
\int_0^1 \left(\Lform{\frac{\delta \ell}{\delta u}}{ \delta u_t} + \Lform{
\frac{\delta \ell}{\delta n}}{\omega_t} + \Lform{\frac{\delta
\ell}{\delta \nu}}{\dot\omega_t - u_t\omega_t -\delta u_t\, n_t}\right) dt =0.
$$
The $\de u$ term yields the equation 
$$\frac{\delta \ell}{\delta u} + \frac{\delta \ell}{\delta \nu} \diamond n_t =
0.$$
(Note the abuse of notation $\de \ell/\de\nu \in T(TN)^*$ is
considered as a linear form on  $TN$ by $\lform{\de \ell/\de\nu}{z} :=
\lform{\de\ell/\de\nu}{(0, z)}$.)
For the $\om$ term, we get, after an integration by parts
$$
\frac{\partial}{\partial t} \frac{\delta
\ell}{\delta \nu} + u_t \star \frac{\delta
\ell}{\delta \nu} - \frac{\delta
\ell}{\delta n} = 0
$$
where we have used the notation 
\begin{equation}
\label{eq:star}
\lform{\frac{\delta
\ell}{\delta \nu}}{u\om} =  \lform{u \star \frac{\delta
\ell}{\delta \nu}}{\om}.
\end{equation}

We therefore obtain the system of equations
\begin{equation}
\label{eq:meta.1}
\left\{
\begin{array}{l}
\displaystyle \frac{\delta \ell}{\delta u} + \frac{\delta \ell}{\delta \nu} \diamond
n_t = 0\\ \\
\displaystyle \frac{\partial}{\partial t} \frac{\delta
\ell}{\delta \nu} + u_t \star \frac{\delta
\ell}{\delta \nu} = \frac{\delta
\ell}{\delta n}\\
\\
\displaystyle
\dot n_t = \nu_t + u_tn_t
\end{array}
\right.
\end{equation}
Note that $\frac{\delta \ell}{\delta u} + \frac{\delta \ell}{\delta \nu} \diamond
n$ is the momentum arising from Noether's theorem for the considered
invariance of the Lagrangian. The special form of the boundary
conditions (fixed $n_0$ and $n_1$) ensures that this momentum is zero.

\section{Euler-Poincar\'e reduction}

An equivalent system can be obtained via an Euler-Poincar\'e
reduction \cite{HoMaRa1998}. In this setting, we make the
variation in the group element and in the template instead of the
velocity and the image. We let $\xi_t  = \de g_t g_t^{-1}$ and
$\varpi_t = g_t\de\eta_t$. From this, we obtain the expressions of $\de u$,
$\de n$ and $\de \nu$. We first have $\de u_t = \dot\xi_t + [\xi_t, u_t]$;
this comes from the standard Euler-Poincar\'e reduction theorem, as
provided in \cite{HoMaRa1998,mr99}. We also have $\de n_t = \de(g_t\eta_t)
= \varpi_t + \xi_t n_t$. 
From $\nu_t = g_t\dot\eta_t$, we get
$\de \nu_t = g_t\de\dot \eta_t + \xi_t\nu_t$ and from $\varpi_t = g_t\de\eta_t$ we also
have $\dot \varpi_t = u_t\varpi_t + g_t\dot\eta_t$. This yields
$\de\nu_t = \dot\varpi_t + \xi_t\nu_t - u_t\varpi_t.$

We also compute the boundary conditions for $\xi$ and $\varpi$. At
$t=0$, we have $g_0 = \id$ and $n_0 = g_0\eta_0 = \text{cst}$ which
implies $\xi_0=0$ and $\varpi_0=0$. At $t=1$, the relation $g_1\eta_1
= \text{cst}$ yields
$\xi_1 n_1 + \om_1 = 0$. 

Now, the first variation is
$$
\int_0^1 \left(\Lform{\frac{\delta \ell}{\delta u}}{\dot\xi_t - \ad_{u_t}\xi_t} + \Lform{
\frac{\delta \ell}{\delta n_t}}{\varpi_t + \xi_t n_t} + \Lform{\frac{\delta
\ell}{\delta \nu}}{\dot\varpi_t + \xi_t\nu_t - u_t\varpi_t}\right) dt =0.
$$
In the integration by parts to eliminate $\dot\xi_t$ and $\dot\varpi_t$,
the boundary term is $\lform{(\de \ell/\de u)_1}{\xi_1} + \lform{(\de\ell
/ \de\nu)_1}{\om_1}$. Using the boundary condition, the last term can
be re-written
$$
- \lform{(\de\ell
/ \de\nu)_1}{\xi_1n_1} = \lform{(\de\ell
/ \de\nu)_1\diamond n_1}{\xi_1}.
$$

We therefore obtain the boundary equation
$$
\frac{\delta \ell}{\delta u}(1) + \frac{\delta \ell}{\delta \nu}(1) \diamond
n_1 = 0.
$$
The evolution equation for $\xi$ is
$$
\frac{\partial}{\partial t} \frac{\delta \ell}{\delta u} + \ad^*_{u_t}
\frac{\delta \ell}{\delta u} + \frac{\delta \ell}{\delta n} \diamond n_t
+ \frac{\delta \ell}{\delta \nu} \diamond \nu_t = 0
$$
and the one for $\varpi$ is
$$
\frac{\partial}{\partial t} \frac{\delta \ell}{\delta \nu} +
u_t \star \frac{\delta \ell}{\delta \nu} - \frac{\delta \ell}{\delta
n} = 0.
$$
We therefore obtain the system
\begin{equation}
\label{eq:meta.2}
\left\{
\begin{array}{l}
\displaystyle
\frac{\partial}{\partial t} \frac{\delta \ell}{\delta u} + \ad^*_{u_t}
\frac{\delta \ell}{\delta u} + \frac{\delta \ell}{\delta n} \diamond n_t
+ \frac{\delta \ell}{\delta \nu} \diamond \nu_t = 0
\\
\\
\displaystyle
\frac{\partial}{\partial t} \frac{\delta \ell}{\delta \nu} + u_t \star \frac{\delta \ell}{\delta \nu} - \frac{\delta \ell}{\delta
n}= 0
\\
\\
\displaystyle
\frac{\delta \ell}{\delta u}(1) + \frac{\delta \ell}{\delta \nu}(1) \diamond
n_1 = 0\\
\\
\displaystyle
\dot n_t = \nu_t + u_tn_t
\end{array}
\right.
\end{equation}

The system \eqref{eq:meta.2} is equivalent to
\eqref{eq:meta.1}, since they characterize the same critical
points. A direct evidence of this fact can be obtained by rewriting
the first equation in \eqref{eq:meta.2} under the form:
$$
\frac{\partial}{\partial t} \Big(\frac{\delta \ell}{\delta u} +
\frac{\delta \ell}{\delta \nu}\diamond u\Big) + \ad_{u_t}^*\Big(\frac{\delta \ell}{\delta u} +
\frac{\delta \ell}{\delta \nu}\diamond u\Big) = 0.
$$

We indeed have, for a solution of \eqref{eq:meta.2},
\begin{eqnarray*}
\frac{\partial}{\partial t}\Big(\frac{\delta \ell}{\delta u_t} + \frac{\delta \ell}{\delta \nu} \diamond
n_t\Big) &=& \frac{\partial}{\partial t} \frac{\delta \ell}{\delta u} +
\Big( \frac{\partial}{\partial t} \frac{\delta \ell}{\delta \nu}\Big)
\diamond n_t + \frac{\delta \ell}{\delta \nu} \diamond \dot n_t\\
&=&
\frac{\partial}{\partial t} \frac{\delta \ell}{\delta u} +
\Big( \frac{\delta \ell}{\delta n} - u_t \star \frac{\delta \ell}{\delta \nu} \Big)
\diamond n_t + \frac{\delta \ell}{\delta \nu} \diamond (\nu_t + u_tn_t)
\\
&=&
\frac{\partial}{\partial t} \frac{\delta \ell}{\delta u} +
\frac{\delta \ell}{\delta n} \diamond n_t + \frac{\delta \ell}{\delta
\nu} \diamond \nu_t -
\Big( u_t\star  \frac{\delta \ell}{\delta \nu} \Big)
\diamond n_t + \frac{\delta \ell}{\delta \nu} \diamond (u_tn_t)
\\
&=&
-\ad^*_{u_t} \frac{\delta \ell}{\delta u} - \ad^*_{u_t}(\frac{\delta \ell}{\delta
\nu} \diamond n_t).
\end{eqnarray*}
In the last equation, we have used the fact that, for any $\al\in\mathfrak g$, 
\begin{eqnarray*}
\Lform{\frac{\delta \ell}{\delta \nu} \diamond (un) - \Big( u\star \frac{\delta \ell}{\delta \nu} \Big)
\diamond n}{\al} &=& \Lform{\frac{\delta \ell}{\delta \nu}}{\al (un) -
u(\al n)} \\
&=& - \Lform{\frac{\delta \ell}{\delta \nu}}{[u, \al] n} \\
&=& - \Lform{\frac{\delta \ell}{\delta \nu} \diamond n}{[u, \al]} \\
&=& \Lform{\ad^*_{u_t}(\frac{\delta \ell}{\delta
\nu} \diamond n_t)}{\al}.
\end{eqnarray*}
This equation combined with $(\delta \ell/\delta u)_1 +
(\delta \ell/\delta \nu)_1\diamond u_1 =0$ obviously implies the first
equation in \eqref{eq:meta.1}.

\section{Special cases}

\subsection{Riemannian metric}
A primary application of this framework can be based on the definition of
a Riemannian metric on $G\times N$ which is invariant for the action
of $G$: $(g, \eta)h = (gh, h^{-1}\eta)$, the corresponding Lagrangian
then taking the form
$$
l(u, n, \nu) = \|(u, \nu)\|^2_n.
$$

The variational problem is now equivalent to the computation of
geodesics for the canonical projection of this metric from $G\times N$
onto $N$. This construction has been introduced in \cite{my01}. The
evolution equations have been derived and studied in \cite{ty05b}
in the case $l(u, n, \nu) = |u|_{\mathfrak g}^2 + |\nu|_n^2$, for a
given norm,  $|.|_{\mathfrak g}$, on $\mathfrak g$ and a pre-existing
Riemannian  structure on $N$.

The interest of this construction is that this provides a Riemannian
metric on $N$ which incorporates the group actions. Examples of this
are given below for point sets and images.

\subsection{Semi-direct product}
Assume that $N$ is a group and that for all $g\in G$, the action of
$g$ on $N$ is a group homomorphism: For all $n, \tilde n\in N$,
$g(n\tilde n) = (gn)(g\tilde n)$ (for example, $N$ can be a vector space and the
action of $G$ can be linear). Consider the semi-direct
product $G \circledS N$ with  $(g,n)(\tilde g ,\tilde n)  =(g\tilde g,
(g\tilde n)n)$ and build on $G\circledS N$ a right-invariant
metric constrained by its value $\|\ \|_{(\id_G,\id_N)}$ at the
identity. Then optimizing the geodesic energy in
$G\circledS N$
between $(\id_G, n_0)$ and $(g_1, n_1)$ with fixed $n_0$ and $n_1$ and
free $g_1$
yields a particular case of
metamorphoses.

Right invariance for the metric on $G\circledS N$ implies
$$
\|(U_g, \ze)\|_{(g,n)} = \|(U\tilde g, (U\tilde n)n + (g\tilde
n)\ze\|_{(g\tilde g, (g\tilde n)n)}
$$
which, using $(\tilde g, \tilde n) = (g^{-1}, g^{-1}n^{-1})$, yields,
letting $u = Ug^{-1}$, 
\begin{eqnarray*}
\|(U_g, \ze)\|_{(g,n)} &=& \|(u, (un^{-1})n + n^{-1}\ze\|_{(\id_G,
\id_N)}\\
&=& \|(u, n^{-1}(\ze - un)\|_{(\id_G,
\id_N)}
\end{eqnarray*}
since $0 = u(n^{-1}n)  = (un^{-1})n +  n^{-1}(un)$. So, the geodesic
energy on $G\circledS N$ for a path of length 1 is
$$
\int_0^1 \|(u_t, n_t^{-1}(\dot n_t - u_tn_t)\|_{(\id_G,
\id_N)}^2
$$
and optimizing this with fixed $n_0$ and $n_1$ is equivalent to
solving the metamorphosis problem with
\begin{equation}
\label{eq:semi.dir}
l(u, n, \nu) =  \|(u, n^{-1}\nu)\|_{(\id_G, \id_N)}^2.
\end{equation}

This turns out to be  a particular case of the previous example. The situation is even
simpler when $N$ is a vector space since this implies $n^{-1}\nu  =
\nu$ for all $n$ and the Lagrangian does not depend on $n$, which
gives a very simple form to systems \eqref{eq:meta.1} and
\eqref{eq:meta.2}. They become
\begin{equation}
\label{eq:meta.1.2}
\left\{
\begin{array}{l}
\displaystyle \frac{\delta \ell}{\delta u} + \frac{\delta \ell}{\delta \nu} \diamond
n_t = 0\\ \\
\displaystyle \frac{\partial}{\partial t} \frac{\delta
\ell}{\delta \nu} + u_t\star  \frac{\delta
\ell}{\delta \nu} = 0\\
\\
\displaystyle
\dot n_t = \nu_t + u_tn_t
\end{array}
\right.
\end{equation}
and
\begin{equation}
\label{eq:meta.2.2}
\left\{
\begin{array}{l}
\displaystyle
\frac{\partial}{\partial t} \frac{\delta \ell}{\delta u} + \ad^*_{u_t}
\frac{\delta \ell}{\delta u} + \frac{\delta \ell}{\delta \nu} \diamond \nu_t = 0
\\
\\
\displaystyle
\frac{\partial}{\partial t} \frac{\delta \ell}{\delta \nu} + u_t\star \frac{\delta \ell}{\delta \nu} = 0
\\
\\
\displaystyle
\frac{\delta \ell}{\delta u}(1) + \frac{\delta \ell}{\delta \nu}(1) \diamond
n_1 = 0\\
\\
\displaystyle
\dot n_t = \nu_t + u_tn_t
\end{array}
\right.
\end{equation}

Even when $N$ is not a vector space, metamorphoses obtained from the
semi-direct product formulation are specific among general
metamorphoses, because they satisfy the conservation of momentum
property which comes with every Lie group with a right invariant
metric. This conservation equation can be written
$$
\Big ( \frac{\delta \ell}{\delta u}, \frac{\delta \ell}{\delta \nu}\Big) =
Ad_{(g_t,n_t)^{-1}}^*\Big ( \frac{\delta \ell}{\delta u}, \frac{\delta
\ell}{\delta \nu}\Big)
$$
where the adjoint representation is the one associated to the
semi-direct product. This property (that we do not
explicit in the general case) will be illustrated in some of the
examples below.

\subsection{Constrained Metamorphoses}

Returning to the general formulation, it is sometimes useful to
include constraints on $n_t$, the image of the metamorphosis, making
the minimization in \eqref{eq:lag} subject to $\Phi(n_t) = 0$ $ (t\in
[0,1])$ for some function $\Phi: N \to \mR^q$. Using Lagrange
multipliers, this directly provides a new version of
\eqref{eq:meta.1}, yielding
\begin{equation}
\label{eq:meta.const}
\left\{
\begin{array}{l}
\displaystyle \frac{\delta \ell}{\delta u} + \frac{\delta \ell}{\delta \nu} \diamond
n_t = 0\\ \\
\displaystyle \frac{\partial}{\partial t} \frac{\delta
\ell}{\delta \nu} + u_t \star \frac{\delta
\ell}{\delta \nu} = \frac{\delta
\ell}{\delta n} - \la_t^T\frac{\de\Phi}{\de n}\\
\\
\displaystyle
\dot n_t = \nu_t + u_tn_t\\ \\
\displaystyle
\lform{\frac{\de\Phi}{\de n}}{\dot n_t} = 0
\end{array}
\right.
\end{equation}

\section{Examples from Pattern Matching}
\label{sec:expl}
In the following examples and in the rest of the paper, $G$ is a group
of diffeomorphisms over some open subset $\Om\subset\mR^d$. We will assume that elements of $G$ can be
obtained as flows associated to ordinary differential equations of the
form $\dot g_t  = u_t \circ g_t$ where $u_t$ is assumed to belong, at
all times, to a Hilbert space $\mathfrak g$ of vector fields on $\Om$ with the condition
$$
\int_0^1 \|u_t\|_{\mathfrak g}^2 dt < \infty.
$$
We will assume that elements of $\mathfrak g$ are smooth enough,
namely that $V$ can be continuously embedded in the space of $C^p$
vector fields with vanishing $p$ first derivatives on $\prt\Om$ and at
infinity, for some $p\geq 1$. More details in this construction can be
found in \cite{ty05} (Appendix C).

We will write the inner product in $\mg$ under the form $\scp{u}{v}_V =
\lform{L_\mg u}{v}$ where $L_\mg$ is the duality operator from $\mg$ to
$\mg^*$. Its inverse, a kernel operator, will be denoted $K_\mg$.

\subsection{Landmarks and Peakons}

The space $N$ contains the objects that are subject to
deformations. The simplest case probably corresponds to configurations
of $Q$ landmarks, for which $N = \Om^Q$. So elements $\eta, \nu\in H$
are $Q$-tuples of points in $\Om$, with tangent vectors being
$Q$-tuples of $d$-dimensional  vectors. 

The model that has been
proposed in \cite{my01, cy01} corresponds to the Lagrangian
$$
\ell(u, n,\nu) = \|u\|_\mg^2 + \frac{1}{\sig^2} \sum_{k=1}^Q |\nu^{(k)}|^2.
$$

This Lagrangian is therefore independent of $n$ (but does not
correspond to a semi direct product). We have $\de \ell/\de u = 2
L_\mg u$, and $(\de \ell/\de\nu) = (2/\sig^2)(\nu^{(1)}, \ldots,
\nu^{(Q)})$. Let $n  = (q^{(1)}, \ldots, q^{(Q)})$. From the definition $\lform{(\de \ell/\de \nu)\diamond n}{w}
= - \lform{\de \ell/\de\nu}{wn}$, we get (since $wn  = (w(q^{(1)}),
\ldots, w(q^{(Q)}))$):
$$
\frac{\de\ell}{\de\nu} \diamond n =  - 2 \sum_{k=1}^Q
\frac{\nu^{(k)}}{\sig^2} \otimes \de_{q^{(k)}}.
$$
Here and later, we use the following notation: if $f$ is a vector
field on $\mR^d$ (considered as a vector density), and $\mu$ a measure
on $\mR^d$, the linear form $f\otimes \mu$, acting on vector fields,
is defined by
\begin{equation}
\label{eq:vec.mes}
\lform{f\otimes\mu}{w}  = \int_{\mR^d} f(x)^Tw(x) d\mu.
\end{equation}

Our first equation for landmark metamorphosis is therefore
$$
L_\mg u_t = \sum_{k=1}^N
\frac{\nu^{(k)}_t}{\sig^2} \otimes \de_{q^{(k)}_t}.
$$
The second equation is $\frac{\partial}{\partial t}\left(\de\ell/\de\nu\right) + u\star (\de\ell/\de
\nu) = 0$, which in this case gives
$$
\dot\nu^{(k)}_t + Du_t(q^{(k)}_t)^T\nu^{(k)}_t = 0, \quad k=1, \ldots, Q.
$$

\begin{figure}
\begin{center}
\includegraphics[width=0.45\textwidth]{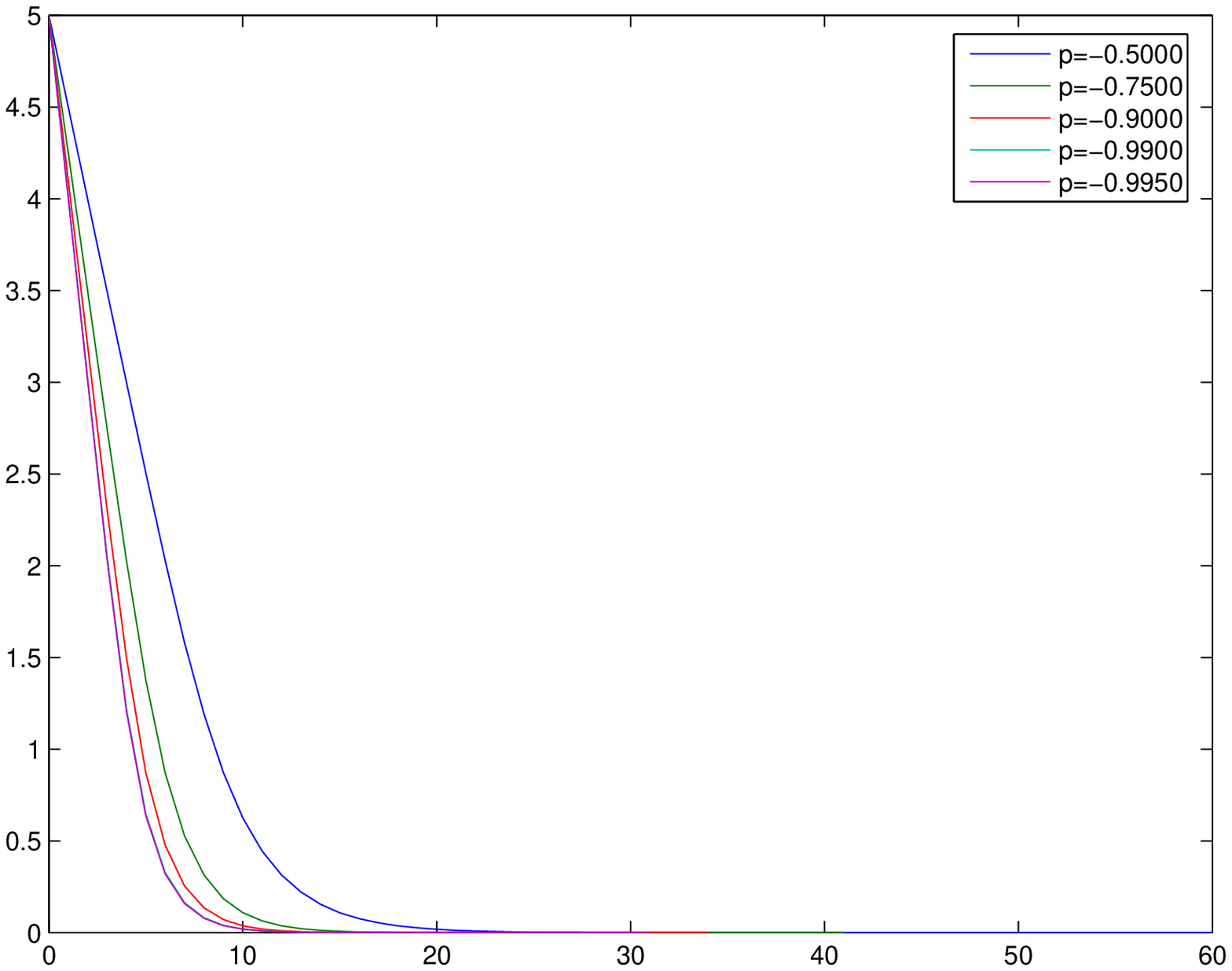}\
\includegraphics[width=0.45\textwidth]{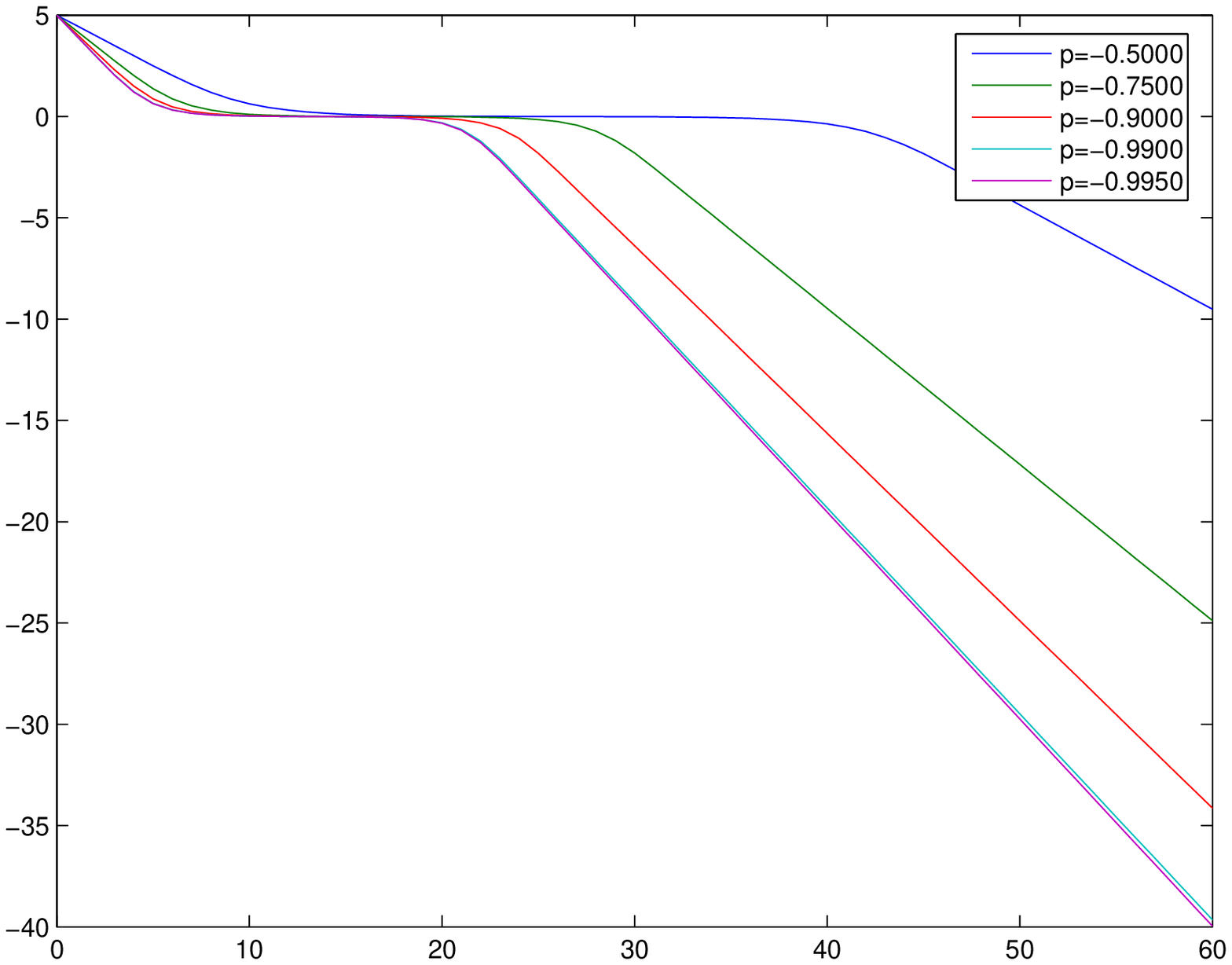}
\caption{\label{fig:peak1} Head-on collision between two peakons. The
plots show the evolution of $q= q_2-q_1$ over time for several values
of $p=p_2-p_1$, for $\sig^2=0$ (left) and $\sig^2 = 10^{-4}$ (right)}
\end{center}
\end{figure}

\begin{figure}
\begin{center}
\includegraphics[width=0.45\textwidth]{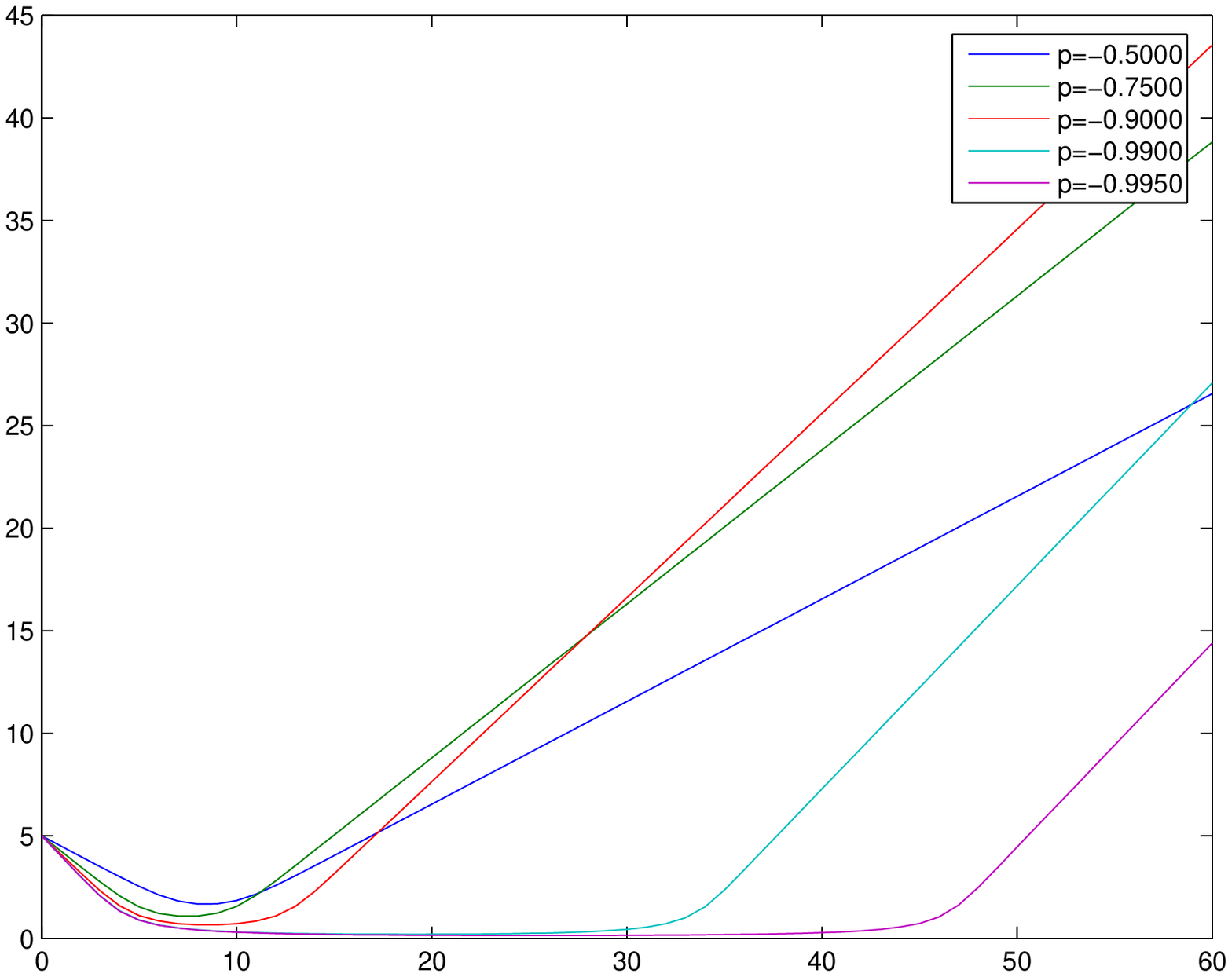}\
\includegraphics[width=0.45\textwidth]{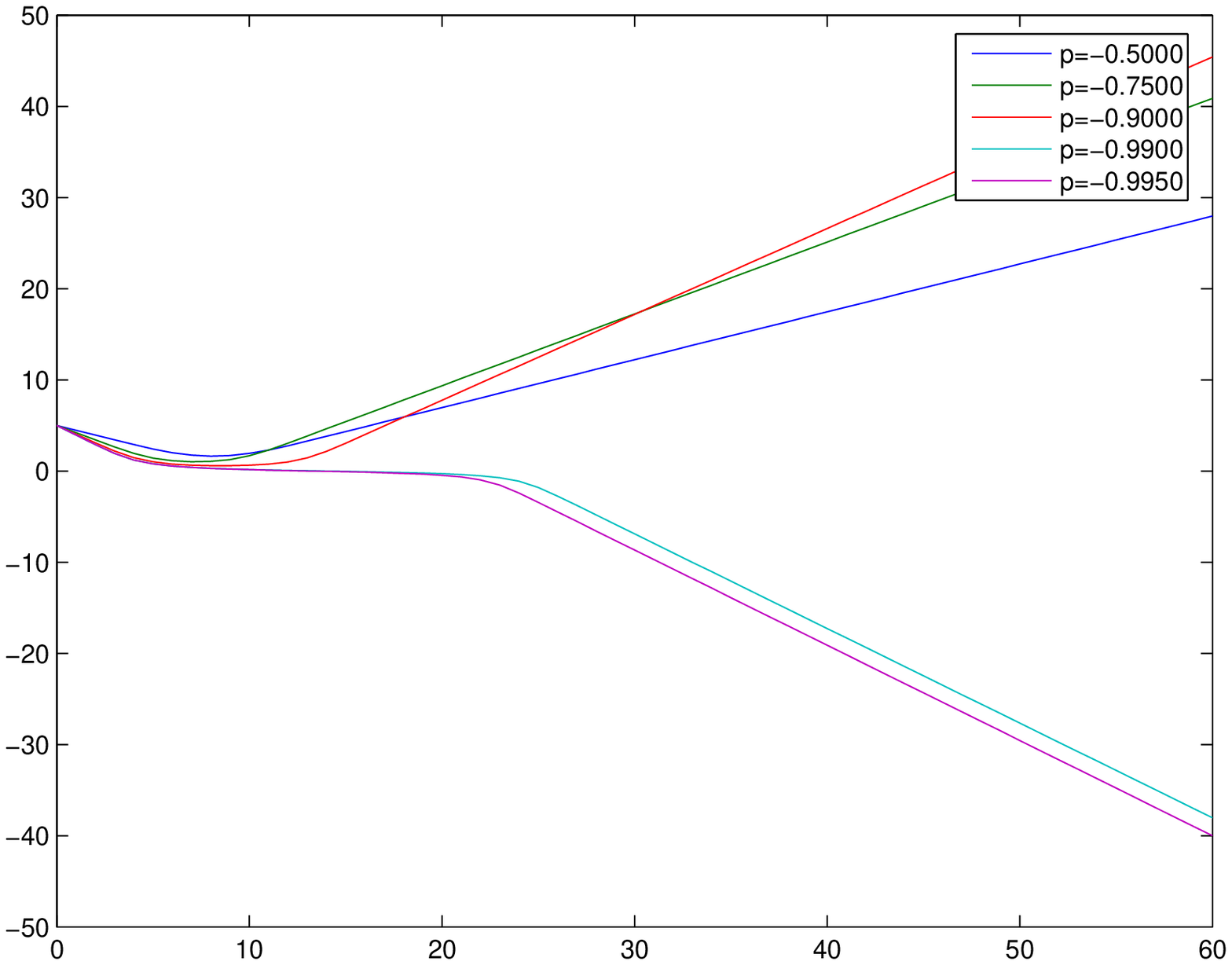}
\caption{\label{fig:peak2} One peakon catching on with another. The
plots show the evolution of $q= q_2-q_1$ over time for several values
of $p=p_2-p_1$, for $\sig^2=0$ (left) and $\sig^2 = 0.05$ (right)}
\end{center}
\end{figure}

Introducing $p^{(k)} = \nu^{(k)}/\sig^2$, we can rewrite system
\eqref{eq:meta.1} in the form:
\begin{equation}
\label{eq:meta.land}
\left\{
\begin{array}{l}
\displaystyle L_\mg u_t = \sum_{k=1}^Q
p^{(k)}_t \otimes \de_{q^{(k)}_t}.
\\ \\
\displaystyle \dot p^{(k)}_t + Du_t(q^{(k)}_t)^Tp^{(k)}_t = 0, \quad k=1, \ldots,
Q\\ \\
\displaystyle
\dot q^{(k)} = u_t(q^{(k)}_t) + \sig^2p^{(k)}_t, \quad k=1, \ldots,
Q
\end{array}
\right.
\end{equation}
Putting the evolution equations into this form is interesting because
the limiting case, $\sig^2 = 0$, exactly corresponds to the peakon
solution of the EPDiff equation \cite{HoMaRa1998}, the dynamics of
which having been recently described in \cite{mcm06}. It is important to
see, however, that the solutions may have significantly distinct
behavior when $\sig >0$. Figures \ref{fig:peak1} and \ref{fig:peak2} illustrate this in
the case of two landmarks in 1D. The plots show the evolution of $r
= q_2 - q_1$ over time when $K_\mg = L_\mg^{-1}$ is a Gaussian kernel. Figure
\ref{fig:peak1} provides a comparison in the case of a head-on
collision ($p_1 + p_2=0$). In the case $\sig^2 = 0$, the peakons
approach each other infinitely closely in time without colliding. For positive $\sig^2$,
the peakons get close, slow down, then cross over  and their distance
grows rapidly to infinity. The duration of the collision phase
($q\simeq 0$) decreases when the sum of the absolute momenta ($p_1-p_2$)
increases. Note that the deformation $g_t$ never becomes singular during
this process. The space first contracts when the peakons get closer,
then expands after the crossover. 

In Figure \ref{fig:peak2}, the case of one peakon overtaking 
another is shown. In the case $\sig^2=0$, we observe a well-known behavior: the peakons approach very closely, then
separate again without crossing over. For metamorphoses 
($\sig^2>0$), the details of the behavior depend on the initial difference between
the momenta. If it is small, then the evolution is similar to the case
$\sig^2=0$. When the initial momentum difference becomes larger, the peakon that
started behind has enough energy to overpass the other one and the two
peakons exchange position. In all cases, the deformations first
experiences a contraction, then an expansion (relative to the position
of the two peakons).

\subsection{Images}
\begin{center}
\begin{figure}
\includegraphics[width=0.15\textwidth]{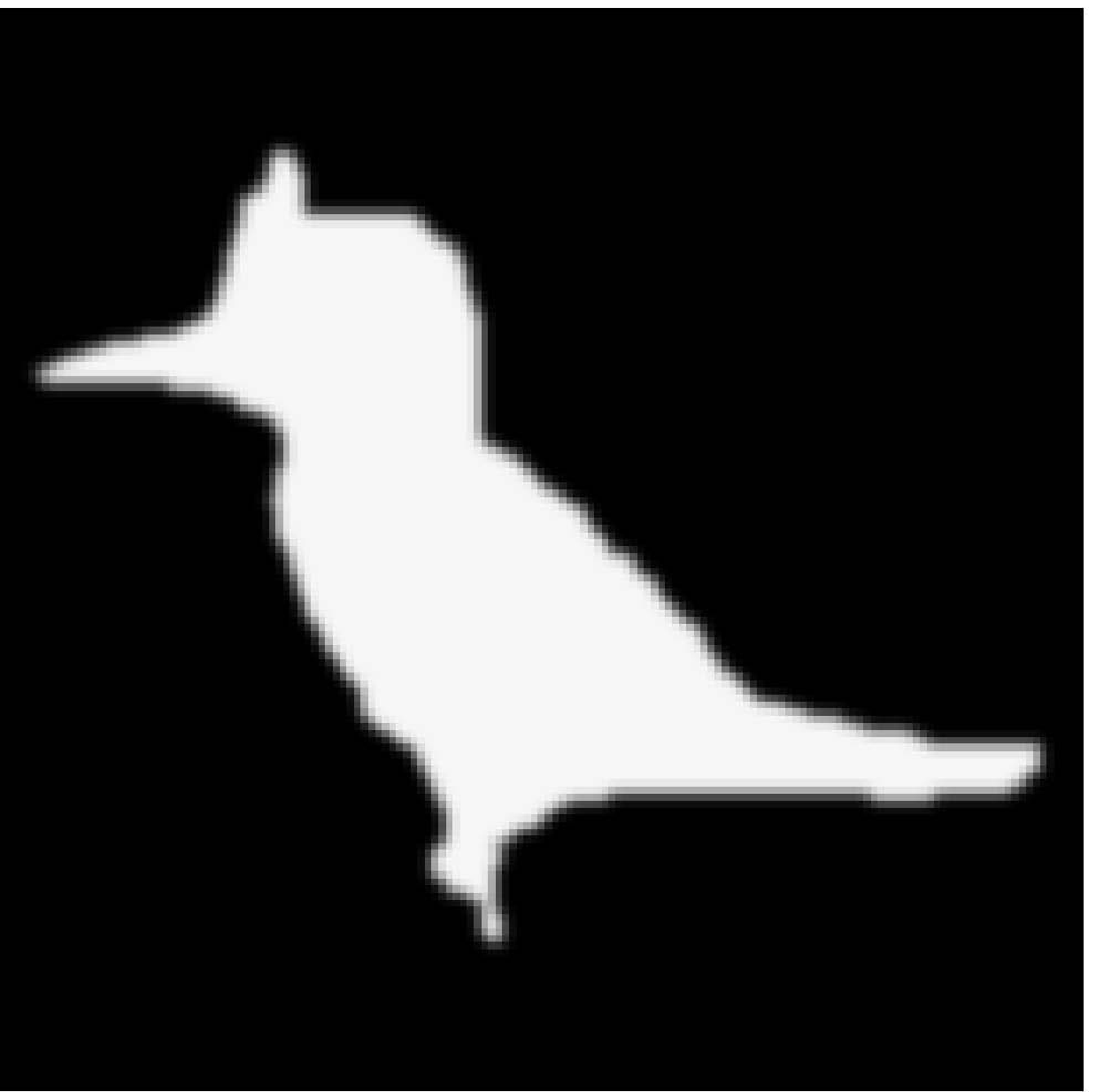} 
\includegraphics[width=0.15\textwidth]{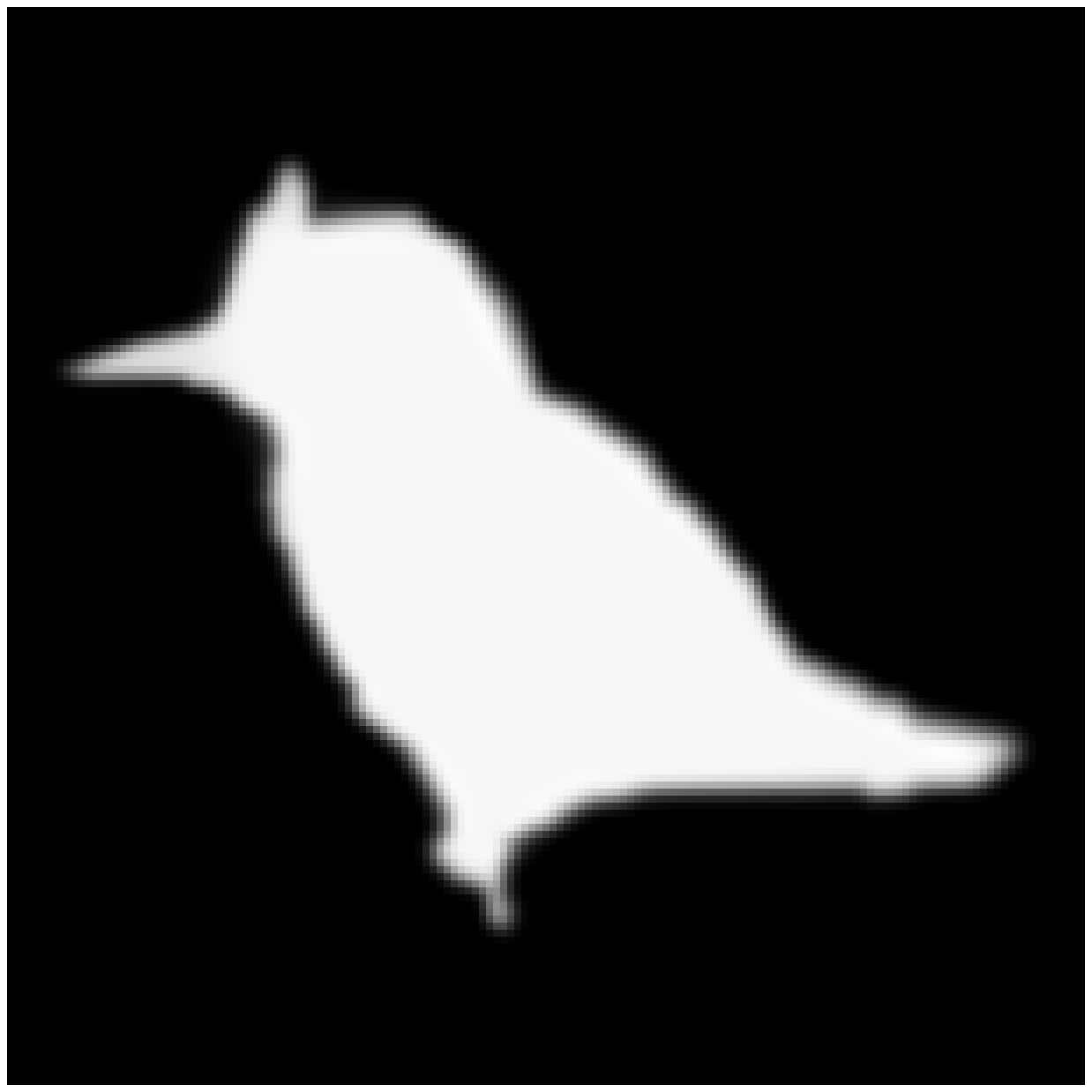} 
\includegraphics[width=0.15\textwidth]{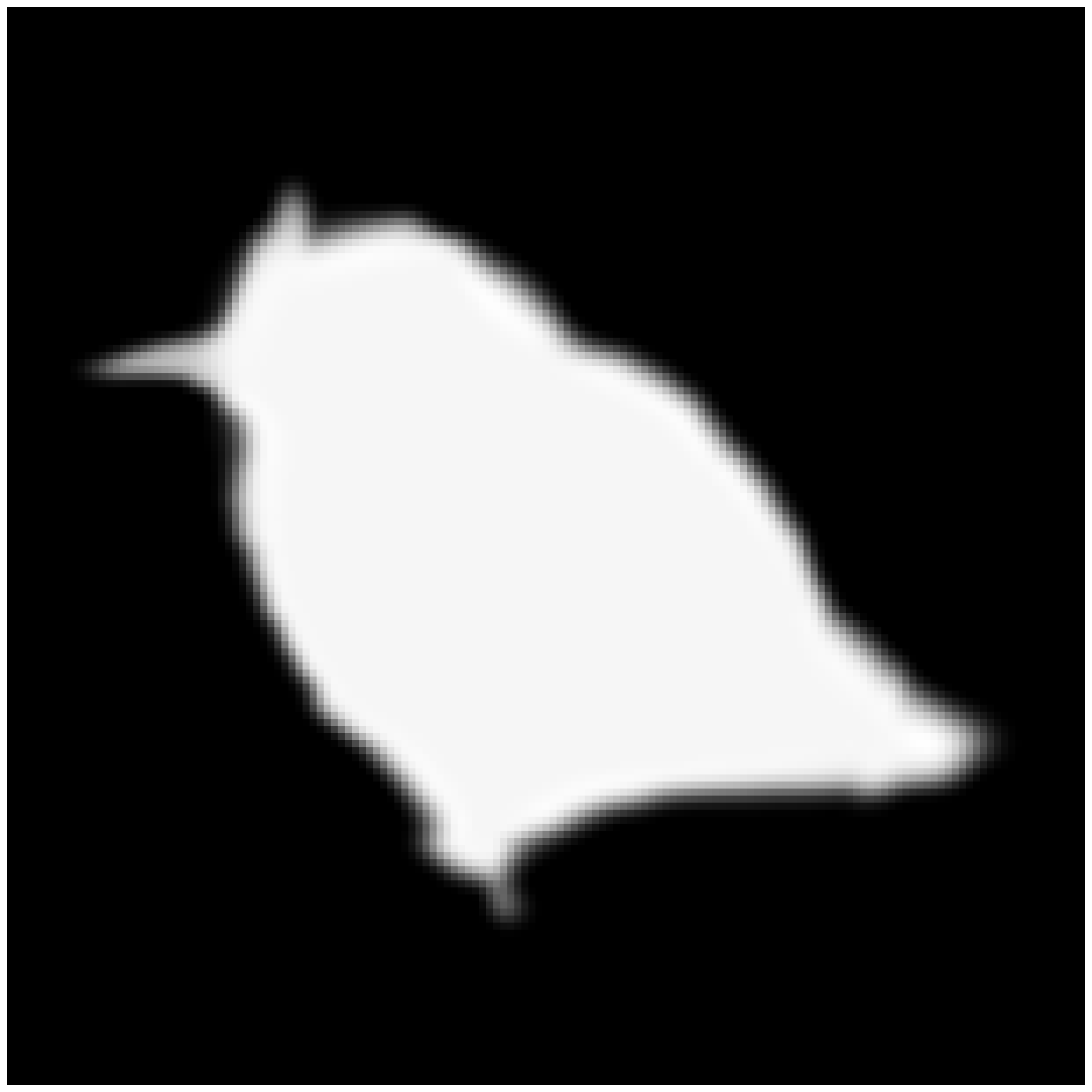} 
\includegraphics[width=0.15\textwidth]{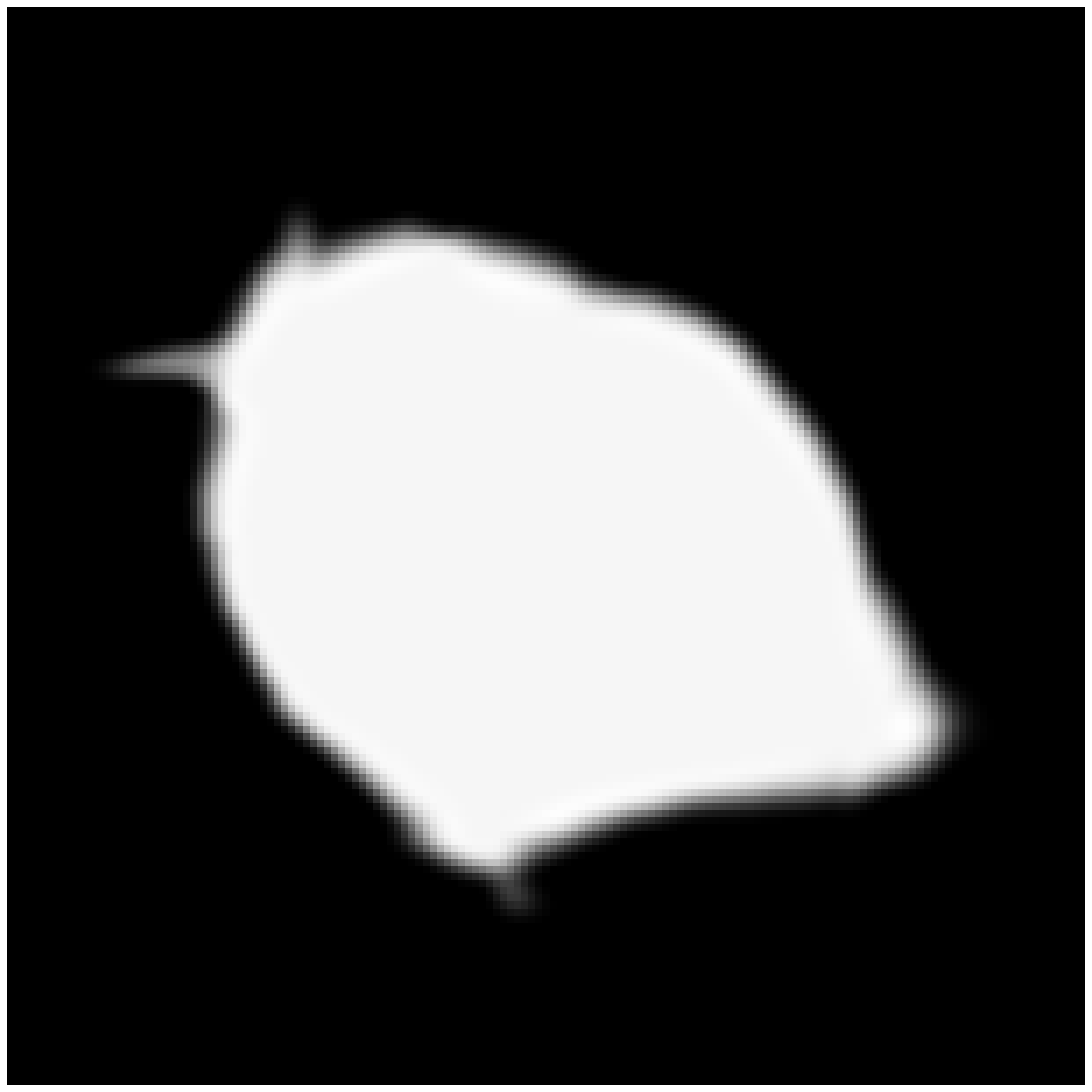} 
\includegraphics[width=0.15\textwidth]{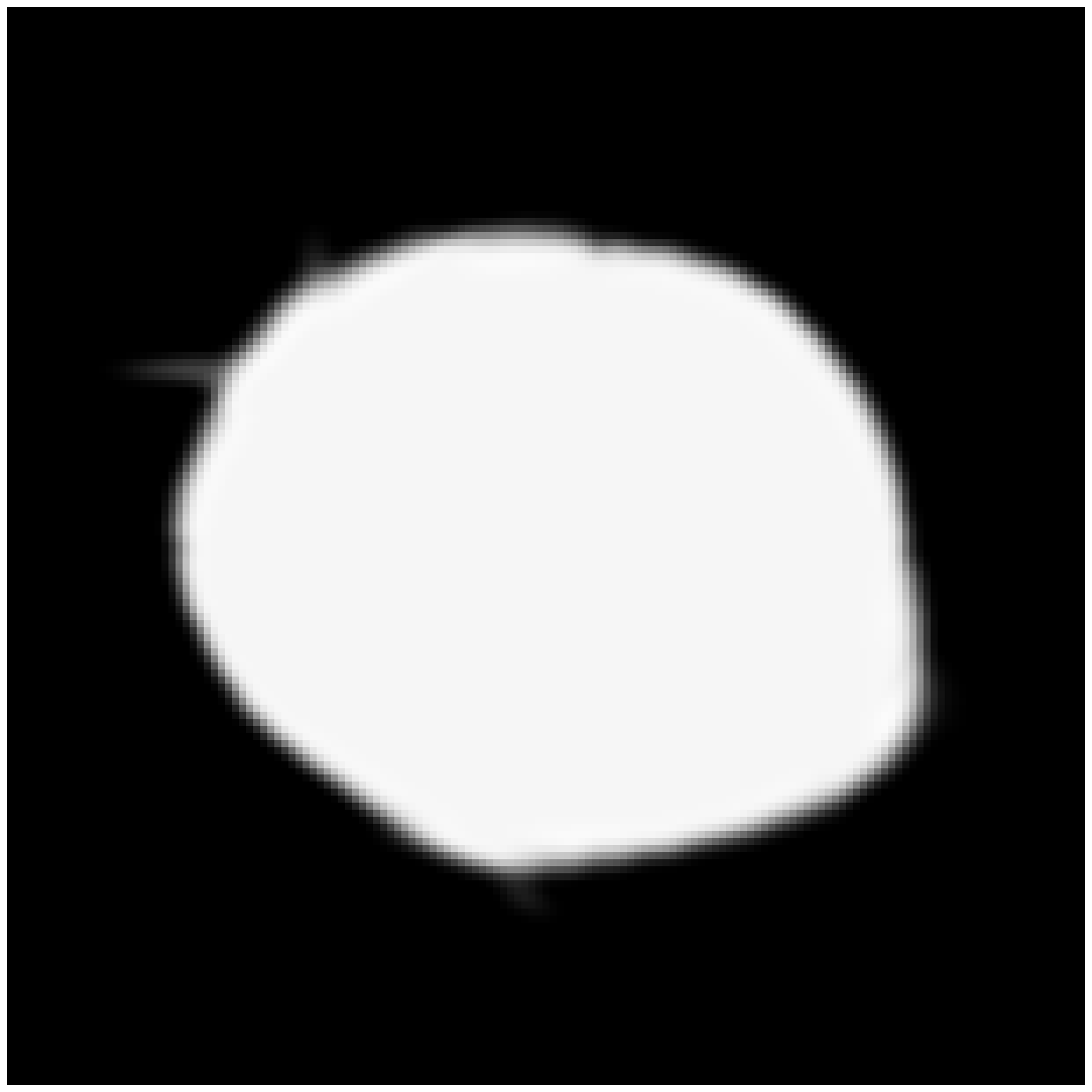} 
\includegraphics[width=0.15\textwidth]{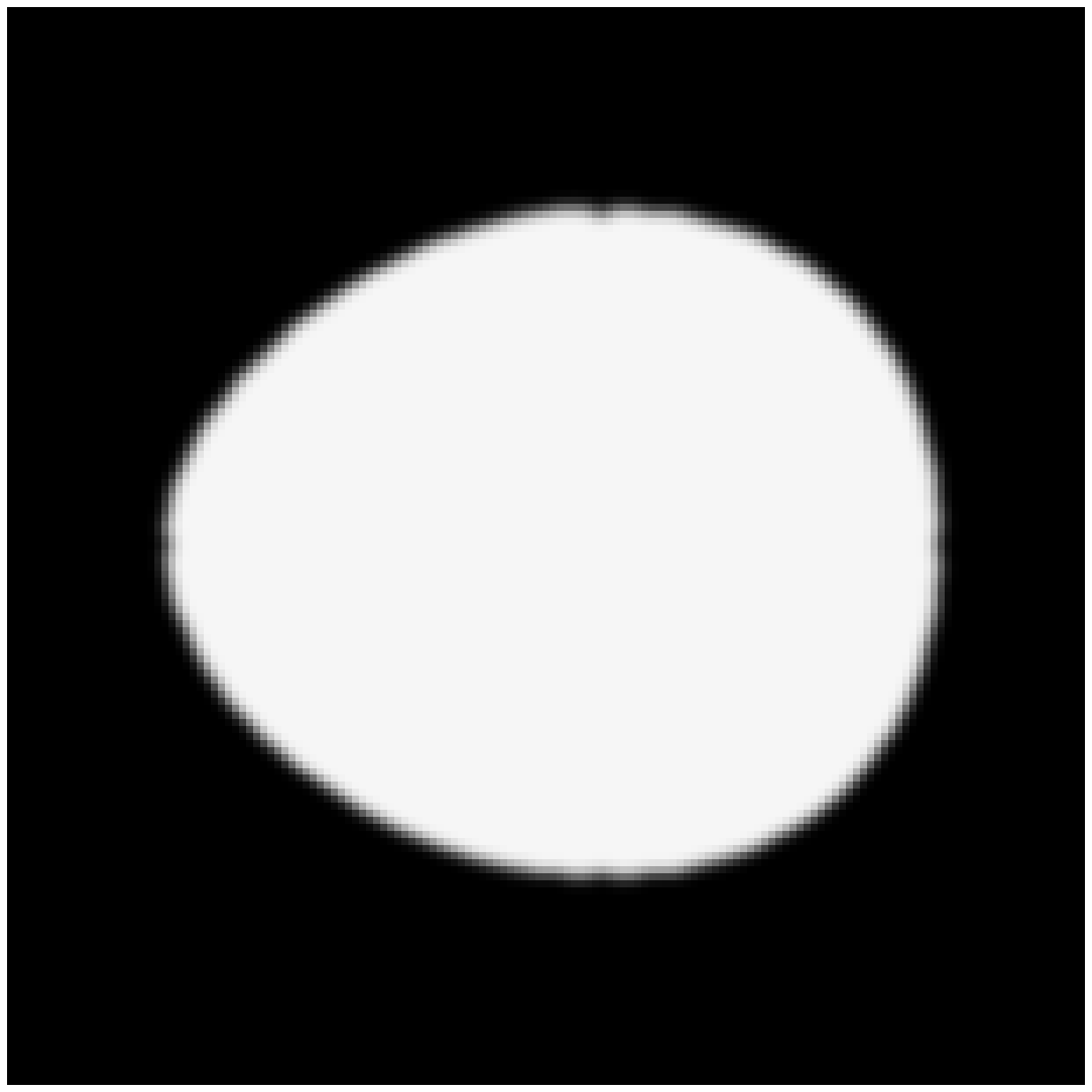} \\
\includegraphics[width=0.15\textwidth]{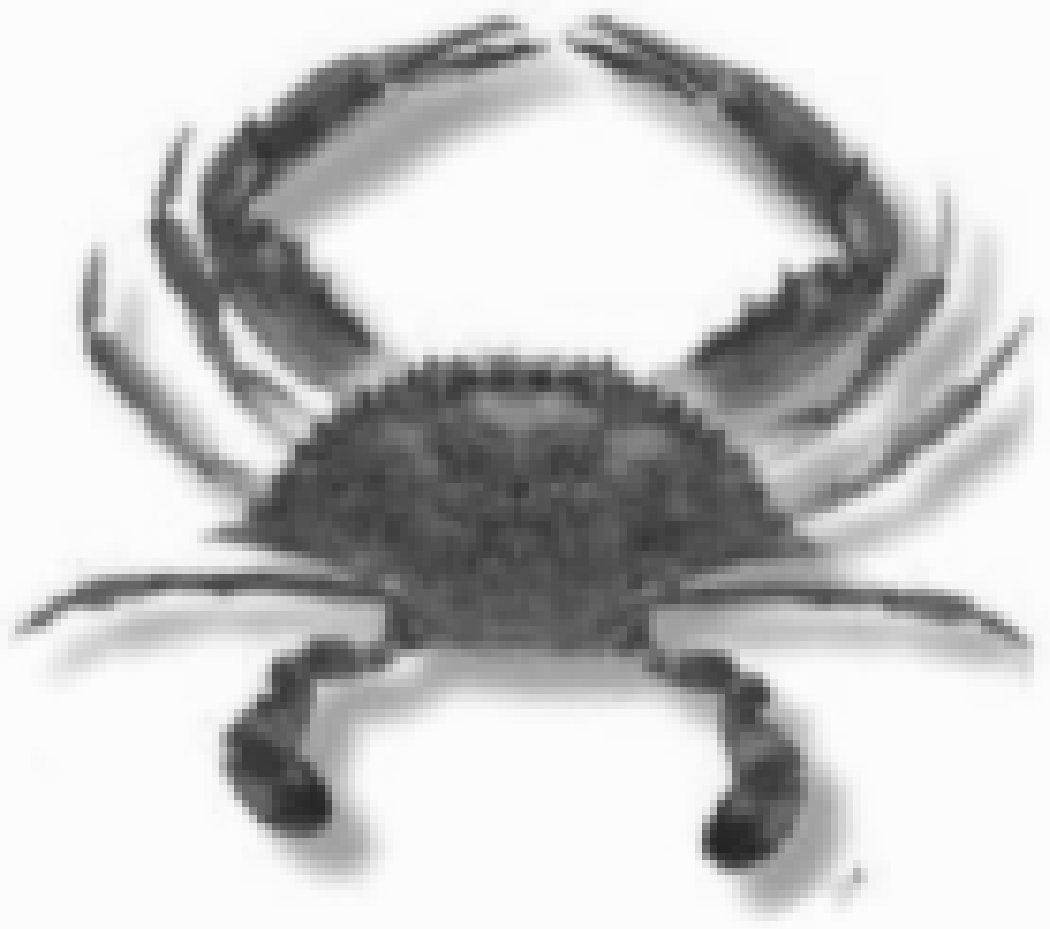} 
\includegraphics[width=0.15\textwidth]{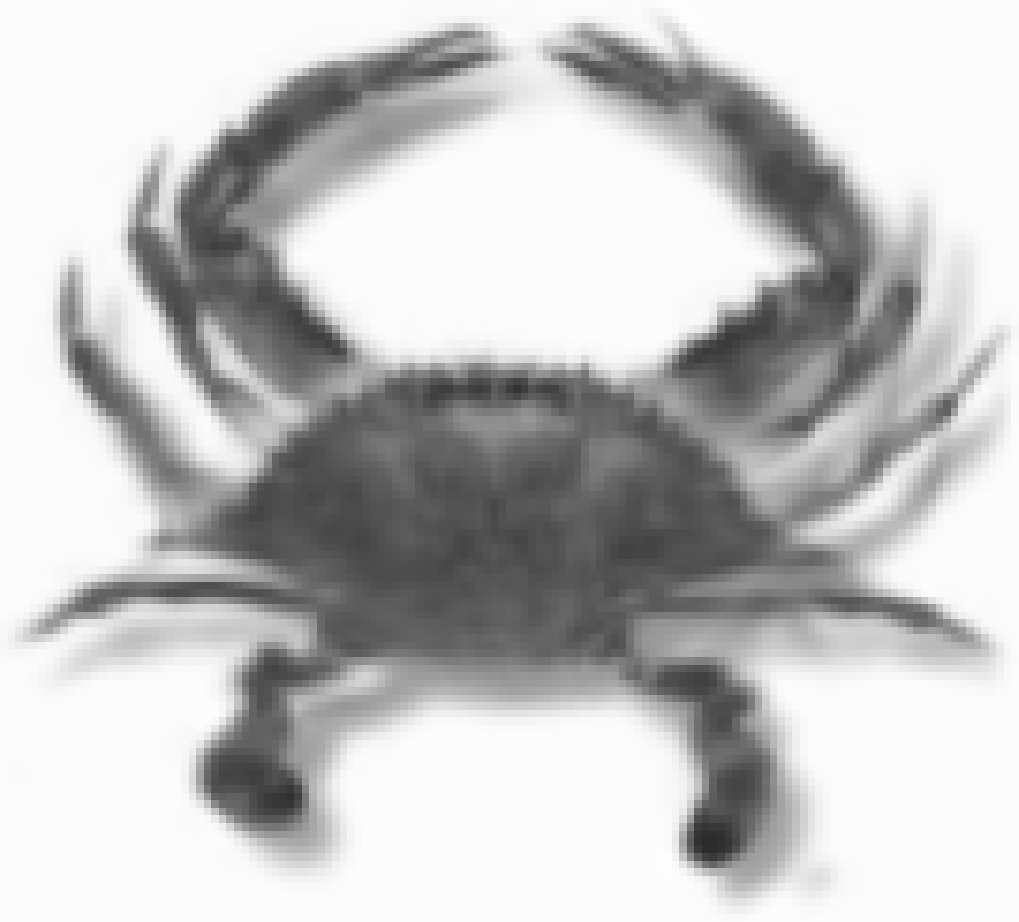} 
\includegraphics[width=0.15\textwidth]{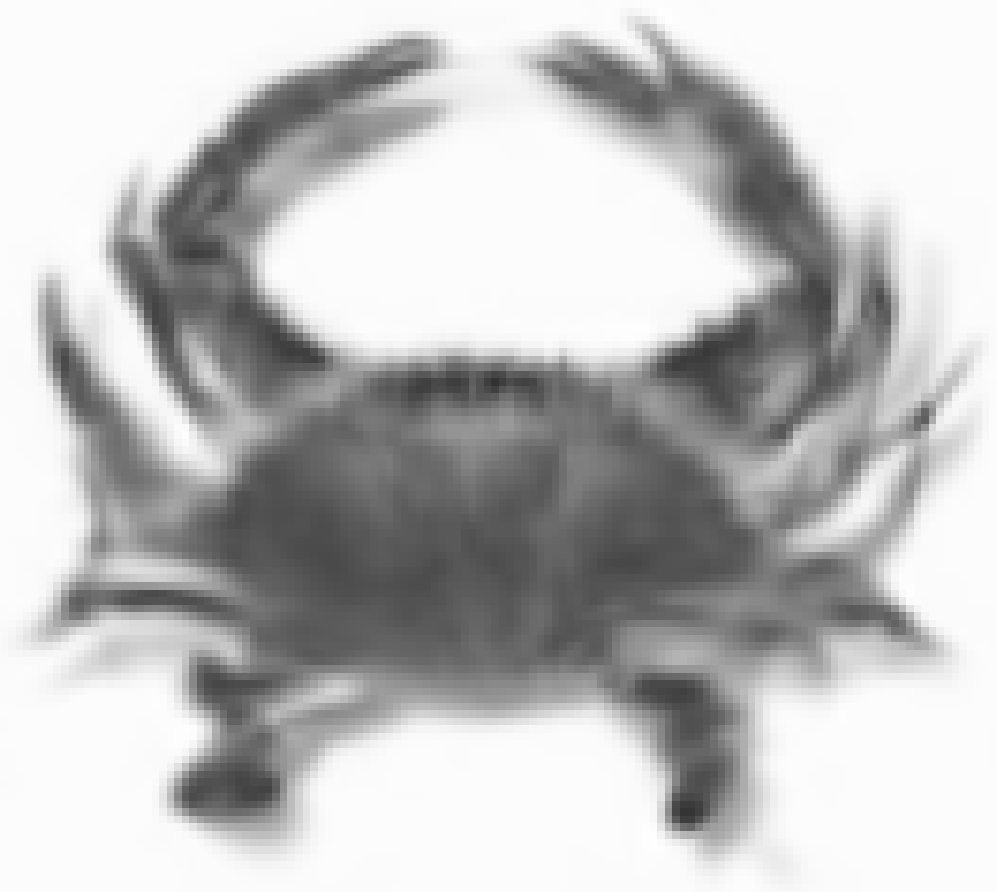} 
\includegraphics[width=0.15\textwidth]{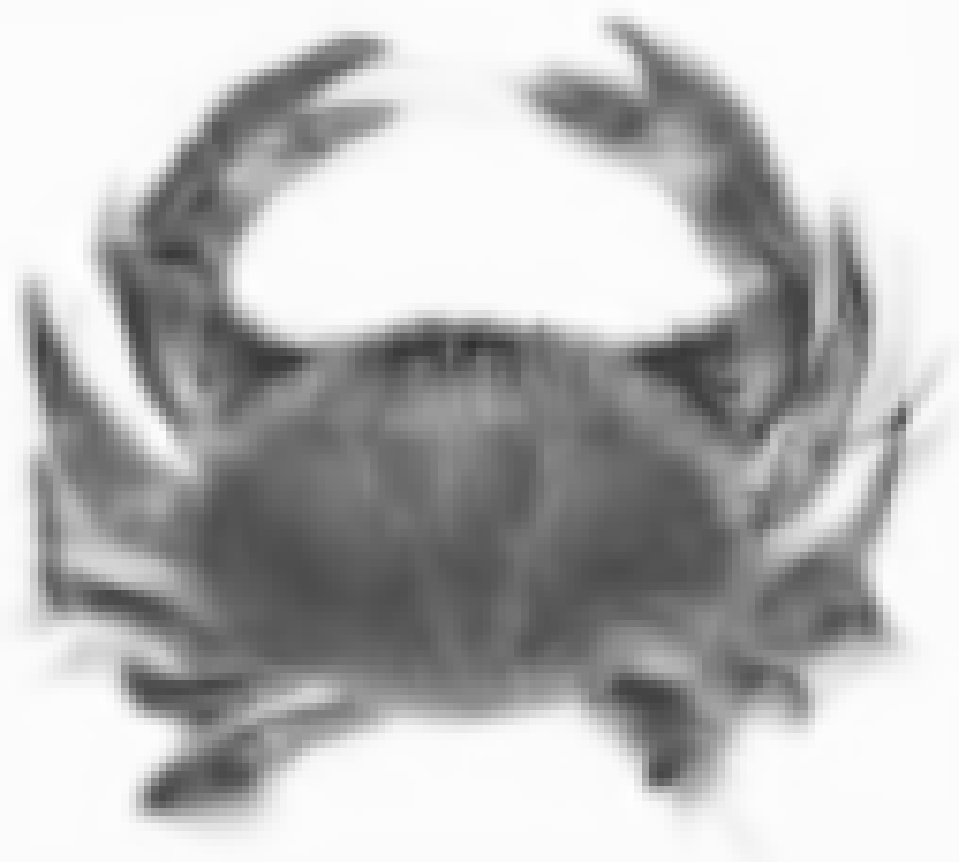} 
\includegraphics[width=0.15\textwidth]{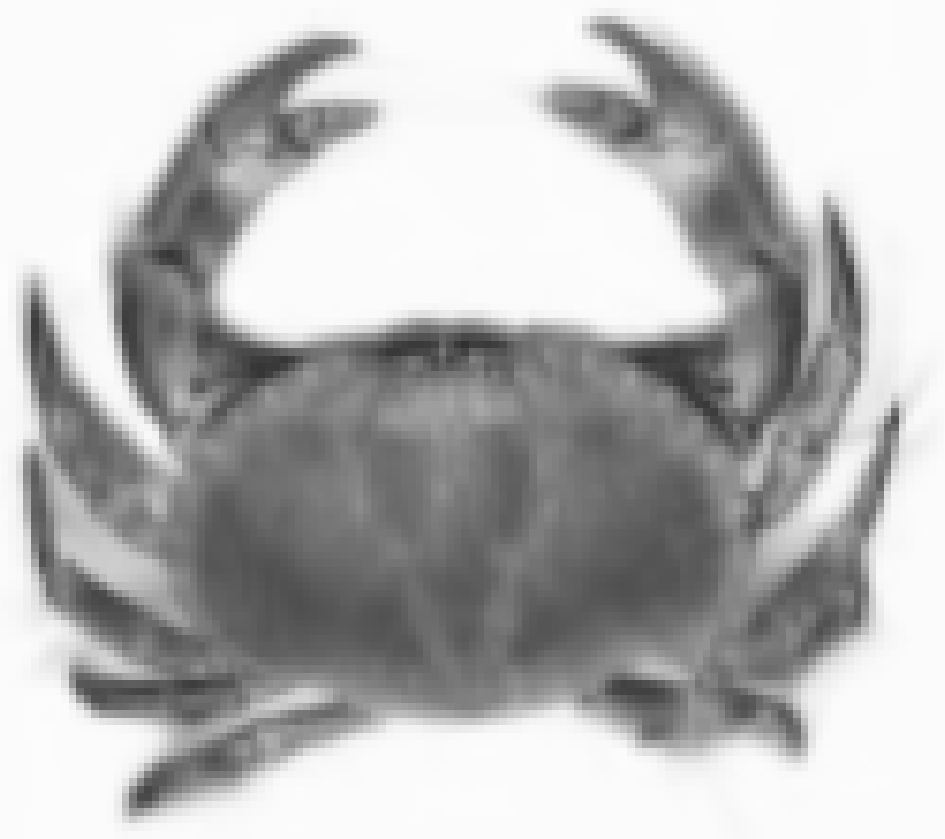} 
\includegraphics[width=0.15\textwidth]{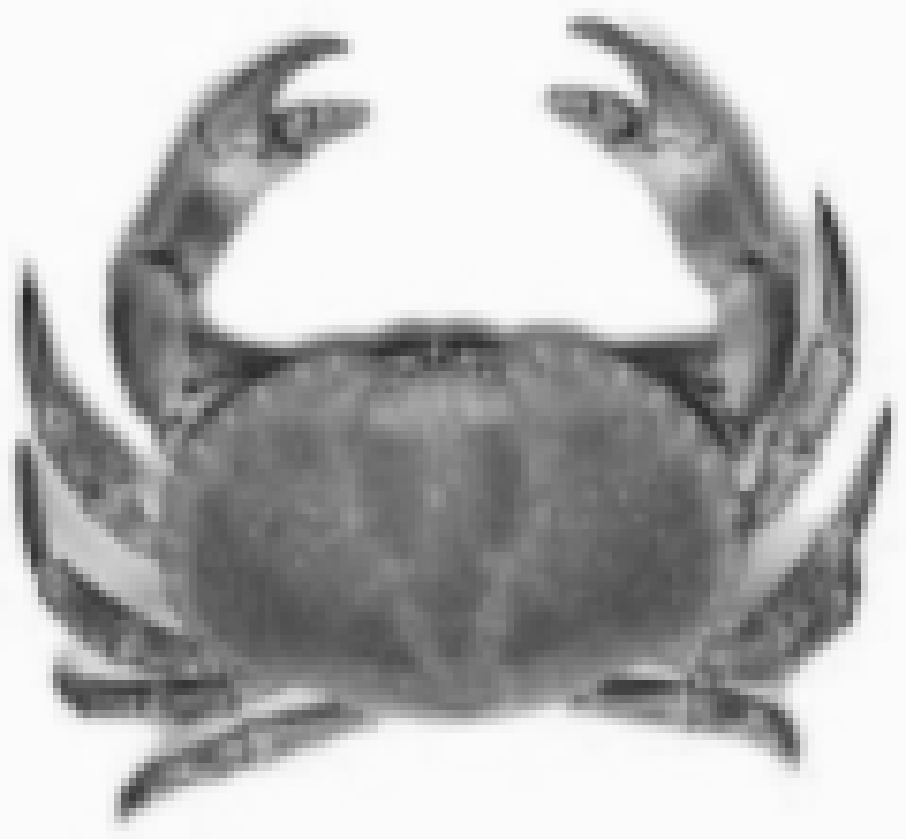} 
\caption{\label{fig:meta.img} Minimizing metamorphosis between
images. The optimal trajectories for $n_t$  are computed
between the first and last images in each row. The remaining images
show $n_t$ at intermediate points in time.}
\end{figure}
\end{center}

Now, consider the case when $N$ is a space of smooth functions from
$\Om$ to $\mR$, that we will call images, with the action $(g, n)
\mapsto n\circ g^{-1}$. A simple case of metamorphoses \cite{my01,ty05} can be
obtained with the Lagrangian
$$
\ell(u, \nu) = \|u\|^2_\mg + \frac{1}{\sig^2} \|\nu\|_{L^2}^2.
$$
If $w\in \mg$ and $n$ is an image, $wn = - \nabla n^T w$, so that
$\lform{\frac{\de \ell}{\de \nu} \diamond n}{w} =
\lform{\frac{\de\ell}{\de\nu}}{\nabla n^T w}$. Thus, since
$\de\ell/\de\nu = 2\nu/\sig^2$, the first equation is
$$
L_\mg u_t = - \frac{1}{\sig^2} \nu_t \nabla n_t.
$$
Since $u\star (\de\ell/\de\nu)$ is defined by
\begin{eqnarray*}
\Lform{u\star \left(\frac{\de \ell}{\de\nu}\right)}{\om} &=&
\Lform{\frac{\de \ell}{\de\nu}}{u\om}\\ 
&=& - \Lform{\frac{\de \ell}{\de\nu}}{\nabla\om^T u}\\ 
&=& - \frac{1}{\sig^2} \lform{\nu}{\nabla\om^T u}\\ 
&=& \frac{1}{\sig^2} \lform{\text{div}(\nu u)}{\om},
\end{eqnarray*}
we obtain the second equation
$$
\dot\nu_t + \frac{1}{\sig^2} \text{div}(\nu_t u_t) = 0.
$$
As in the landmark case, denote $z = \nu/\sig^2$ and rewrite the
evolution equations in the form
\begin{equation}
\label{eq:meta.img}
\left\{
\begin{array}{l}
\displaystyle L_\mg u_t = -  z_t\nabla n_t \\
\\
\displaystyle \dot z_t + \text{div}(z_tu_t) = 0\\ \\
\displaystyle
\dot n_t + \nabla n_t^Tu_t = \sig^2 z_t
\end{array}
\right.
\end{equation}
Existence and uniqueness of solutions for this system have been proved
in \cite{ty05}. From a visual point of view, image metamorphoses
are similar to what is usually called ``morphing'' in computer
graphics. The evolution of the image over time, $t\mapsto n_t$, is a
combination of deformations and image intensity variation. Algorithms
and experimental results 
for the solution of the boundary value problem (minimize the Lagrangian
between two images) can be found in \cite{my01, gy05}. Some examples
of minimizing geodesics are also provided in Figure
\ref{fig:meta.img}.

In 1D, letting $m=L_{\mg}u=(1-\partial_x^2)u$, the time evolving form
of this system (as provided by \eqref{eq:meta.2}, or by direct
computation from \eqref{eq:meta.dens}) becomes, with $\rho = \sig z$:
\begin{eqnarray}
&&\hspace{5mm}
\partial_t m+u\partial_x m + 2m\partial_xu = - \rho\partial_x\rho
\quad\hbox{with}\quad
\partial_t\rho + \partial_x(\rho u) = 0
\label{SDPsystem1D}
\end{eqnarray}
This relates, with the important difference of a minus sign in front
of $\rho\partial_x\rho$ in the first equation, to the two-component Camassa-Holm system studied in \cite{clz05,Falqui06,Ku2007}. The system (\ref{SDPsystem1D}) in our case is equivalent to 
the compatibility for $d\lambda/dt=0$ of 
\begin{eqnarray}
\partial_x^2\psi 
&+& \Big( -\,\frac{1}{4} 
+ m\lambda + \rho^2\lambda^2\Big )\psi = 0
\label{LaxPair-eigen}
\\
\partial_t\psi
 &=& 
-\,\Big( \frac{1}{2\lambda} + u\Big)\partial_x\psi
+ \frac{1}{2}\psi\partial_x u
\label{LaxPair-evol}
\end{eqnarray}


Image matching can also be seen under the semi-direct product point of
view, since the action is linear and the Lagrangian takes the form
\eqref{eq:semi.dir} with $n^{-1}\nu = \nu$. This implies that the
momentum, which is, in this case, the pair $(L_\mg u, z)$, is conserved in a fixed
frame. Working out the conservation equation $\Ad_{(g, n)}^*(L_\mg u, z)
= \cst$ in this case yields the equations $L_\mg u_t + z_t\nabla n_t = \mathrm{cst}$
and $z_t =\mathrm{det}(Dg_t^{-1}) z_0 \circ g_t^{-1}$. This last
condition is the integrated form of the second equation in
\eqref{eq:meta.img}, while the first equation in \eqref{eq:meta.img}
implies that in fact $L_\mg u_t + z_t\nabla n_t=0$, which is the
horizontality condition in the quotient space $G\circledS N/G$.

\subsection{Densities}
\label{sec:dens}
We here let $N$ be a space of smooth functions $n:\Om\to\mR$ with the
action $(g, n) \mapsto |\det D(g^{-1})| n\circ g^{-1}$, i.e., $n$
deforms as a density. We consider the same Lagrangian as with images, 
$$
\ell(u, \nu) = \frac{1}{2}\|u\|^2_\mg + \frac{1}{2\sig^2} \|\nu\|_{L^2}^2.
$$
For $w\in\mg$ and $n\in H$, we have $wn = - \nabla n^T w -
n\text{div}(w) = -\text{div}(nw)$. This implies
\begin{eqnarray*}
\lform{\frac{\de\ell}{\de\nu}\diamond n}{w} &=& - \frac{1}{\sig^2}
\lform{\nu}{\text{div}(nw)}\\
&=& \frac{1}{\sig^2} \lform{n\nabla\nu}{w}
\end{eqnarray*}
yielding the first equation
$$
L_\mg u = \frac{1}{\sig^2} n\nabla\nu.
$$
Similarly, we get $u \star \nu = \nabla \nu^T u$ and the equation
$$
\dot\nu +\nabla\nu^T\nu = 0.
$$
This yields the system, where we have, as before, introduced $z=\nu/\sig^2$:
\begin{equation}
\label{eq:meta.dens}
\left\{
\begin{array}{l}
\displaystyle L_\mg u =  n\nabla z \\
\\
\displaystyle \dot z + \nabla z^T u = 0\\ \\
\displaystyle
\dot n + \text{div}(nu) = \sig^2 z
\end{array}
\right.
\end{equation}

\begin{center}
\begin{figure}
\includegraphics[width=0.15\textwidth]{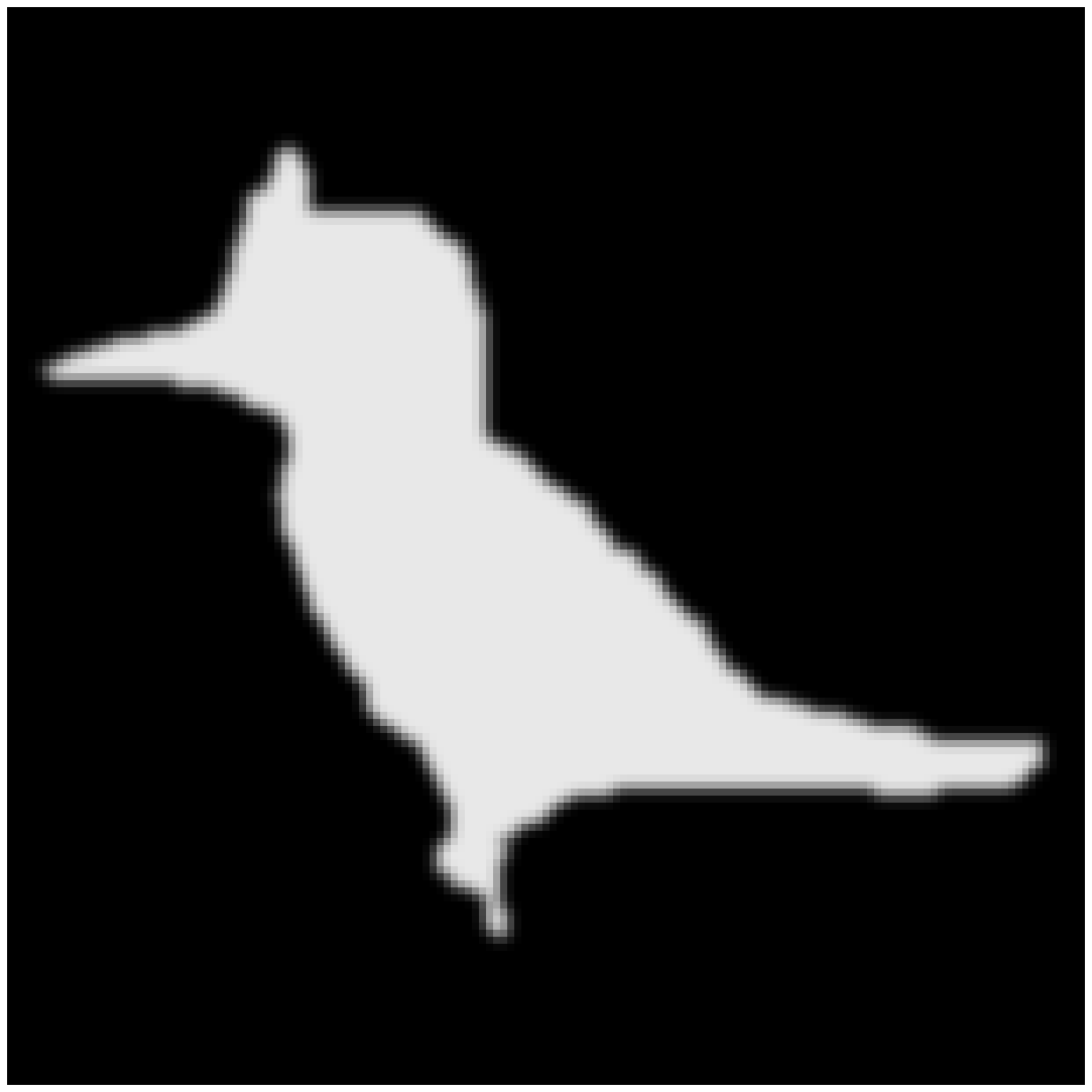} 
\includegraphics[width=0.15\textwidth]{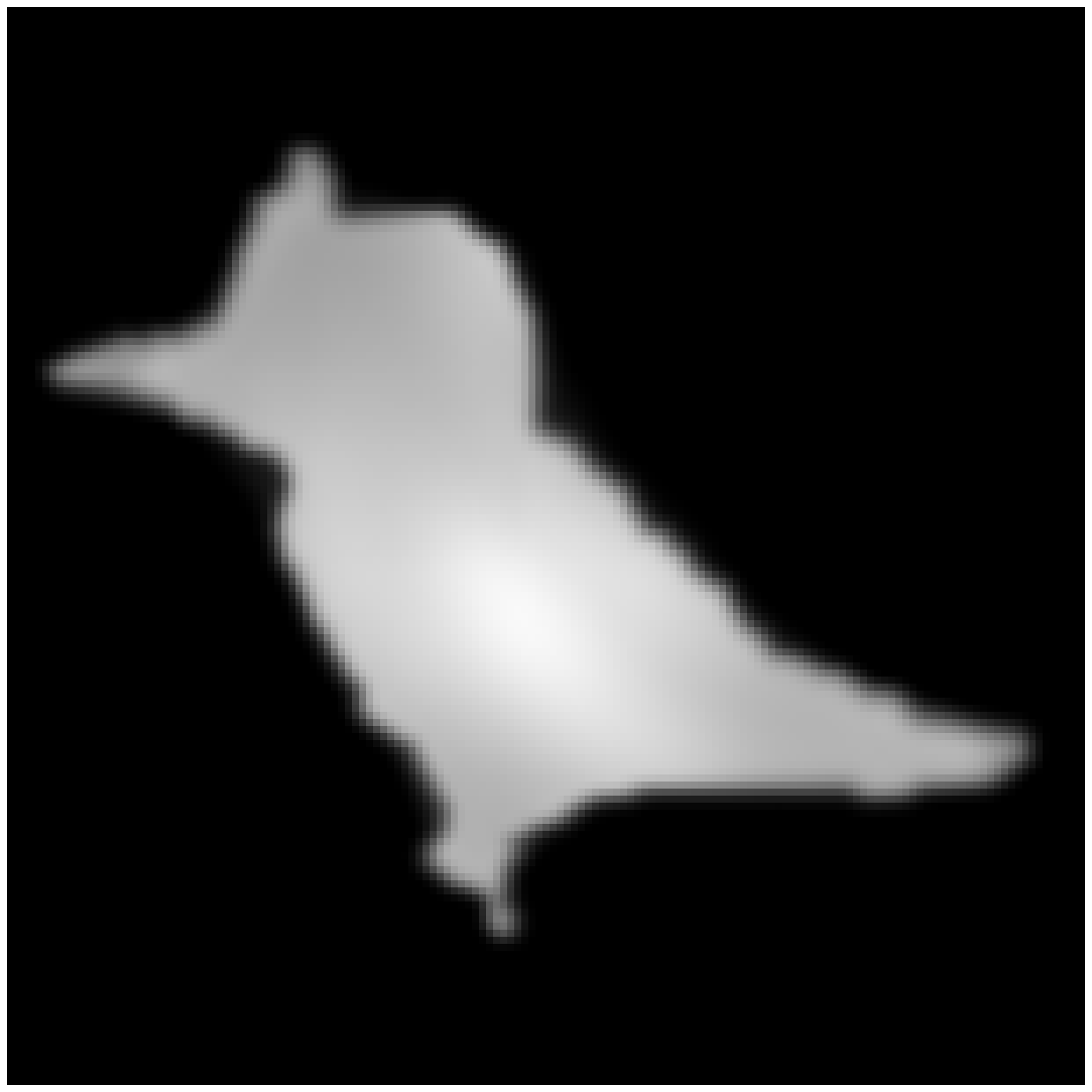} 
\includegraphics[width=0.15\textwidth]{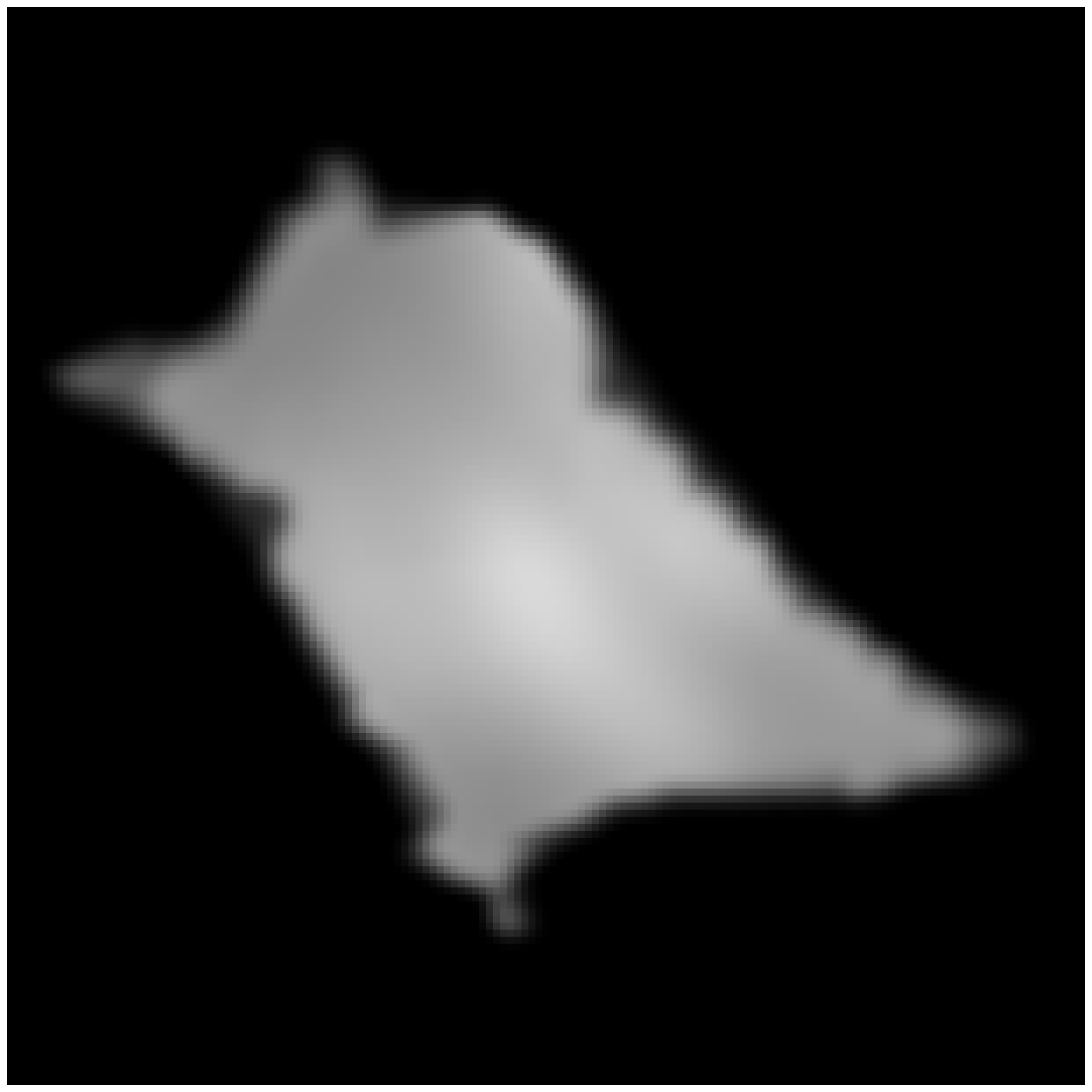} 
\includegraphics[width=0.15\textwidth]{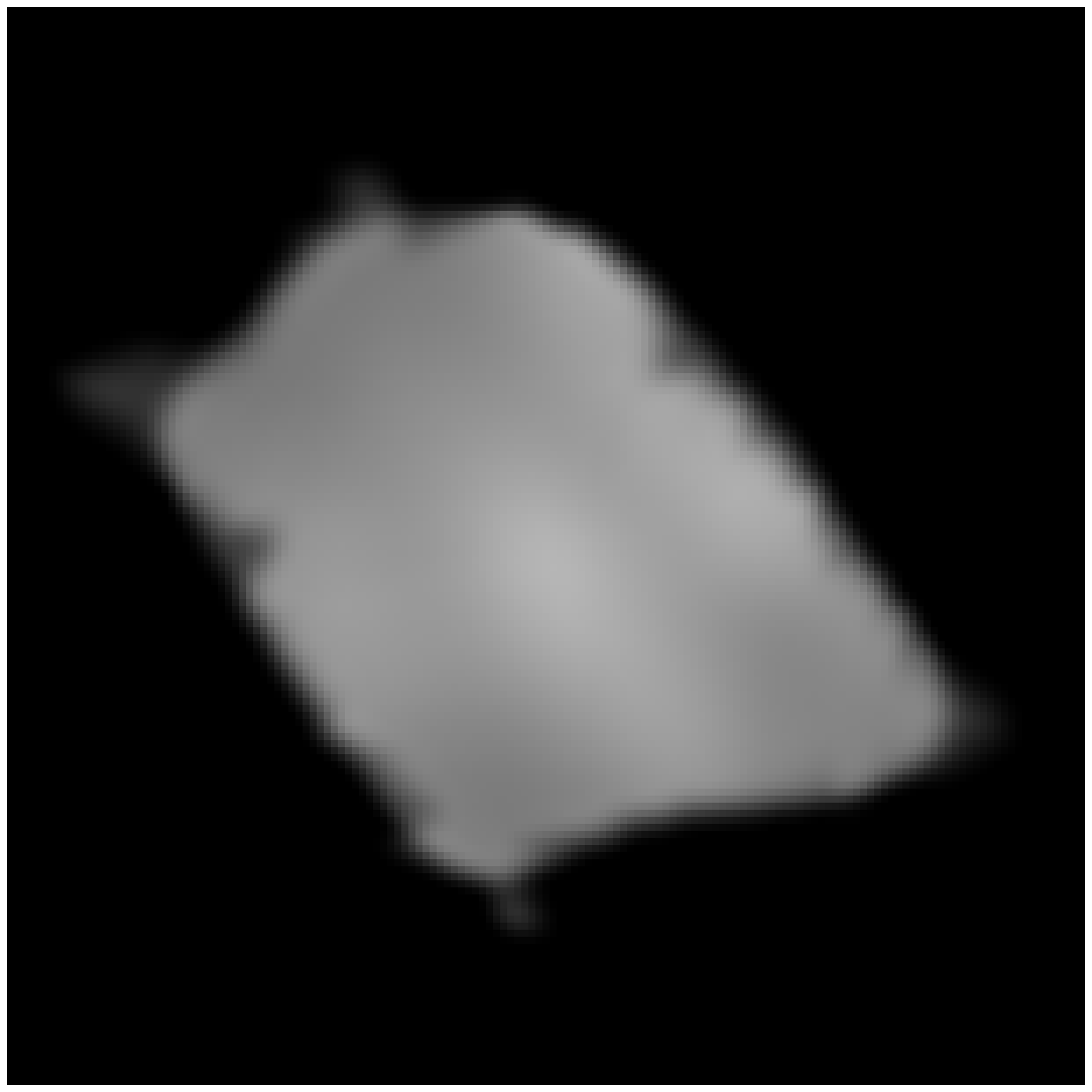} 
\includegraphics[width=0.15\textwidth]{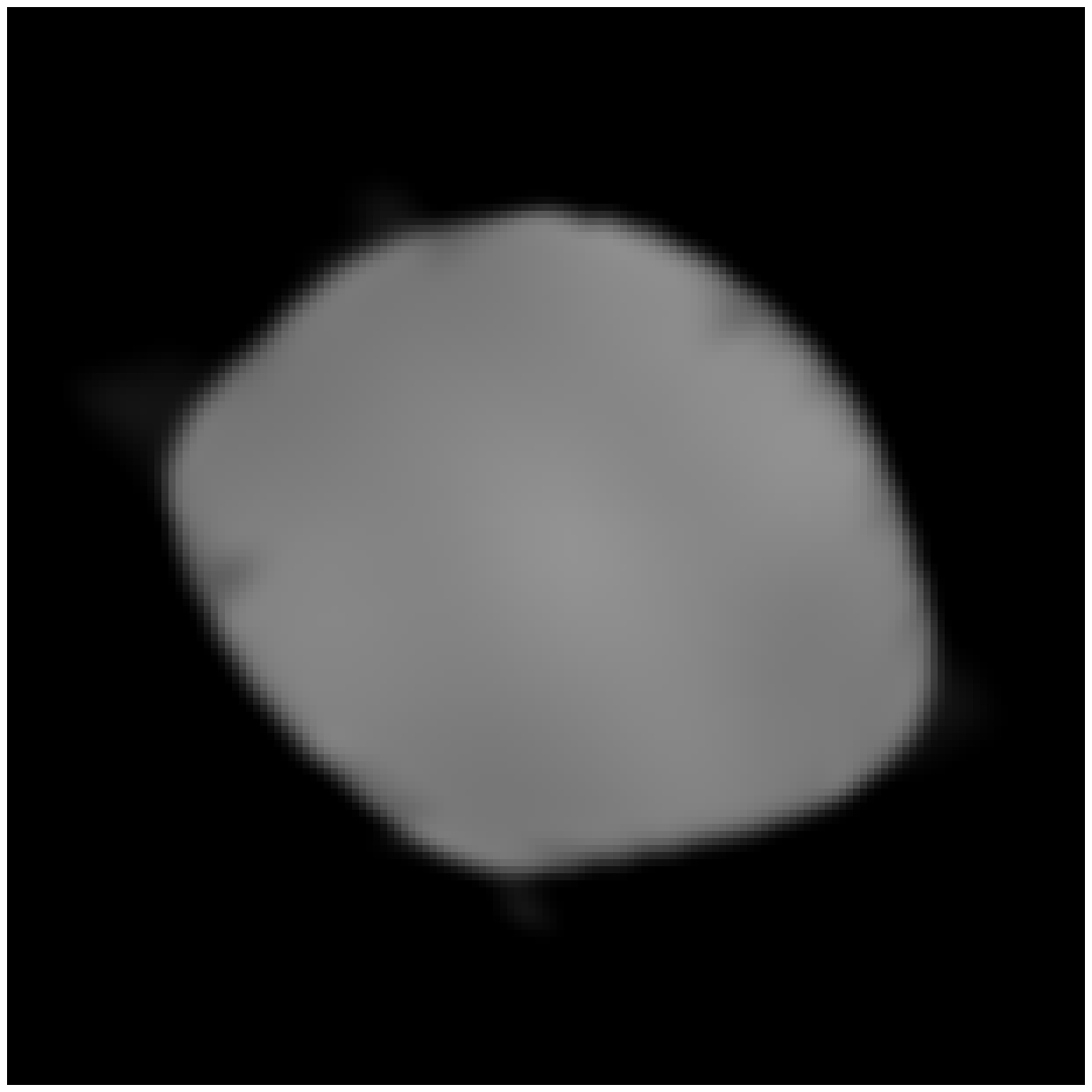} 
\includegraphics[width=0.15\textwidth]{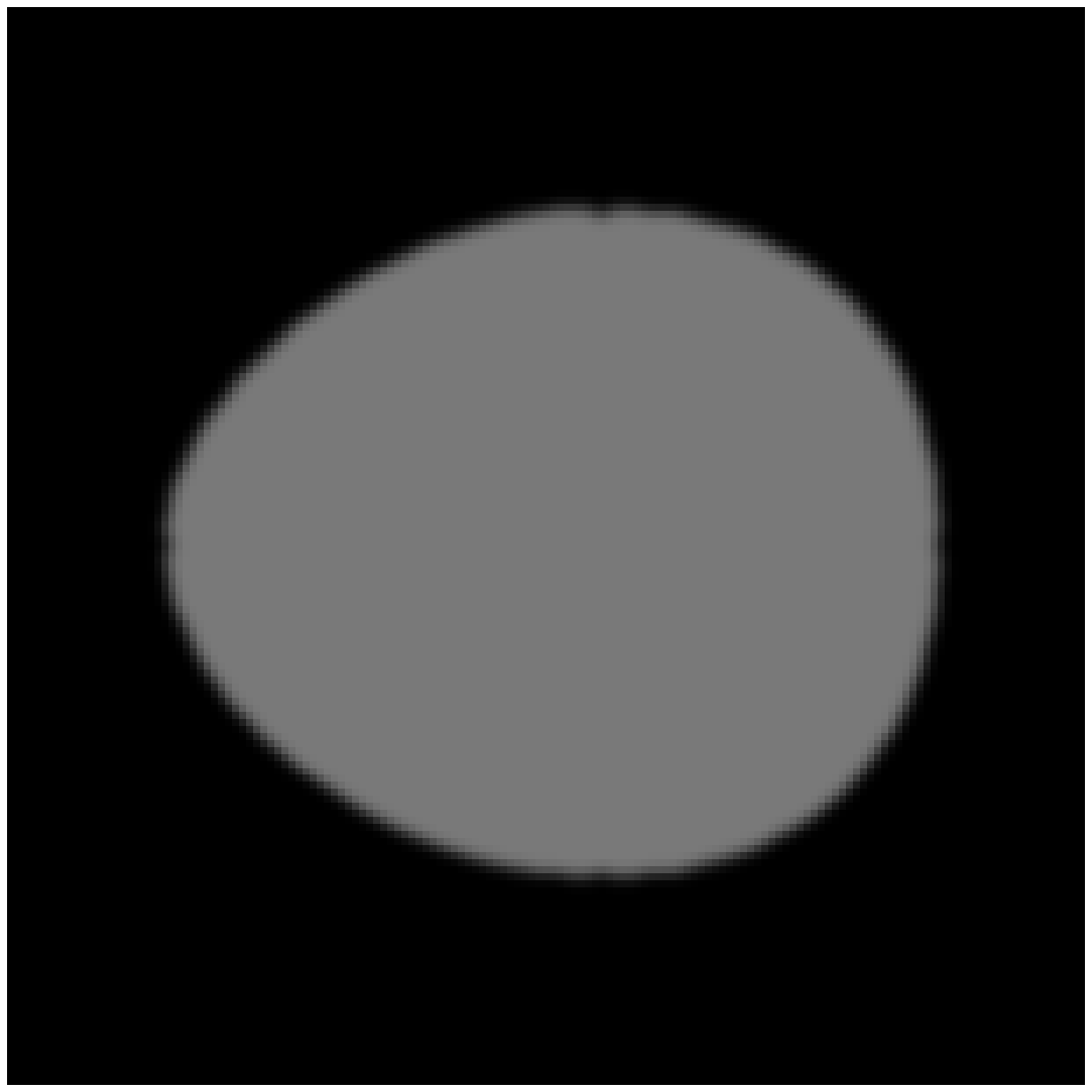} \\
\includegraphics[width=0.15\textwidth]{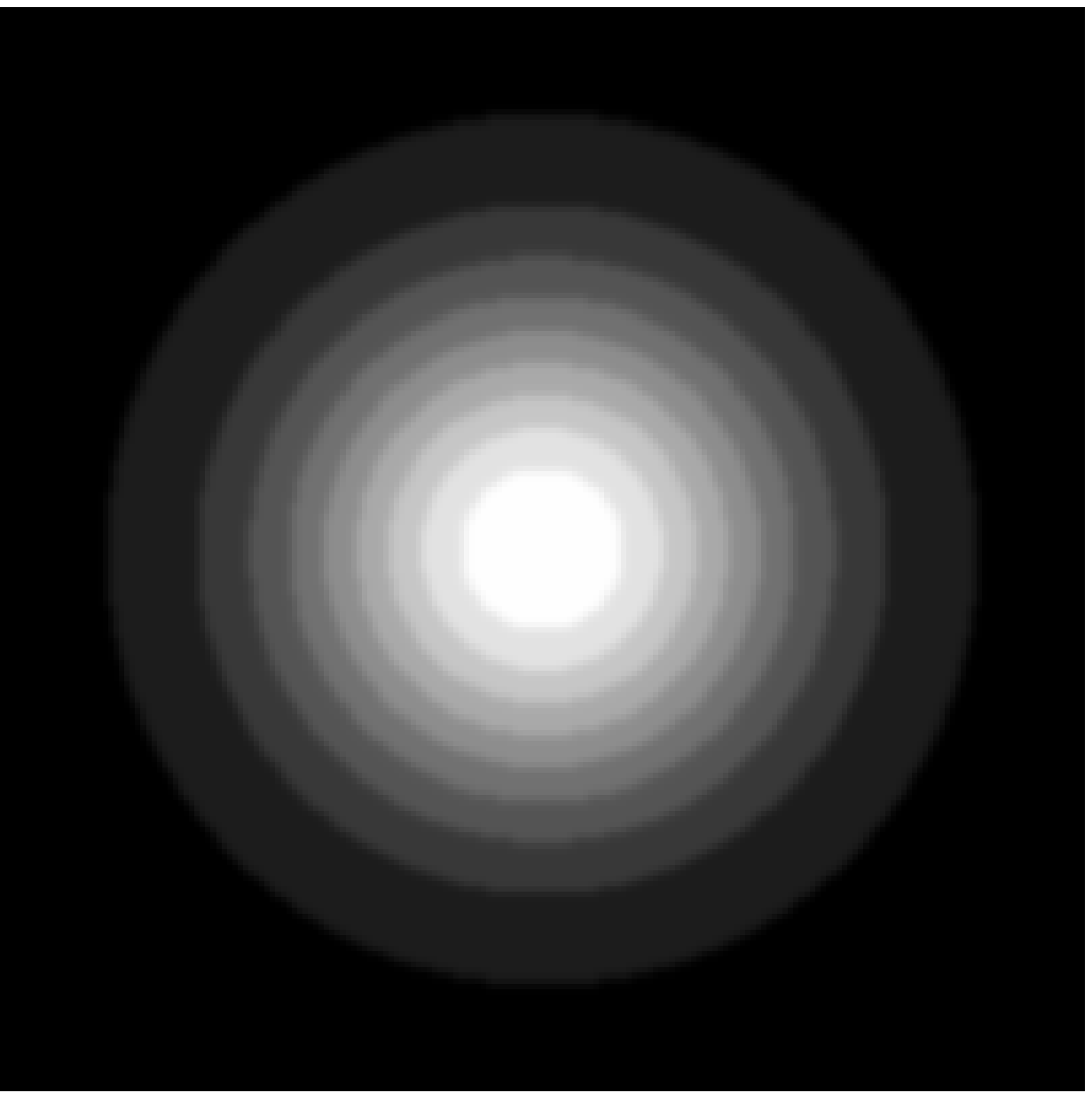} 
\includegraphics[width=0.15\textwidth]{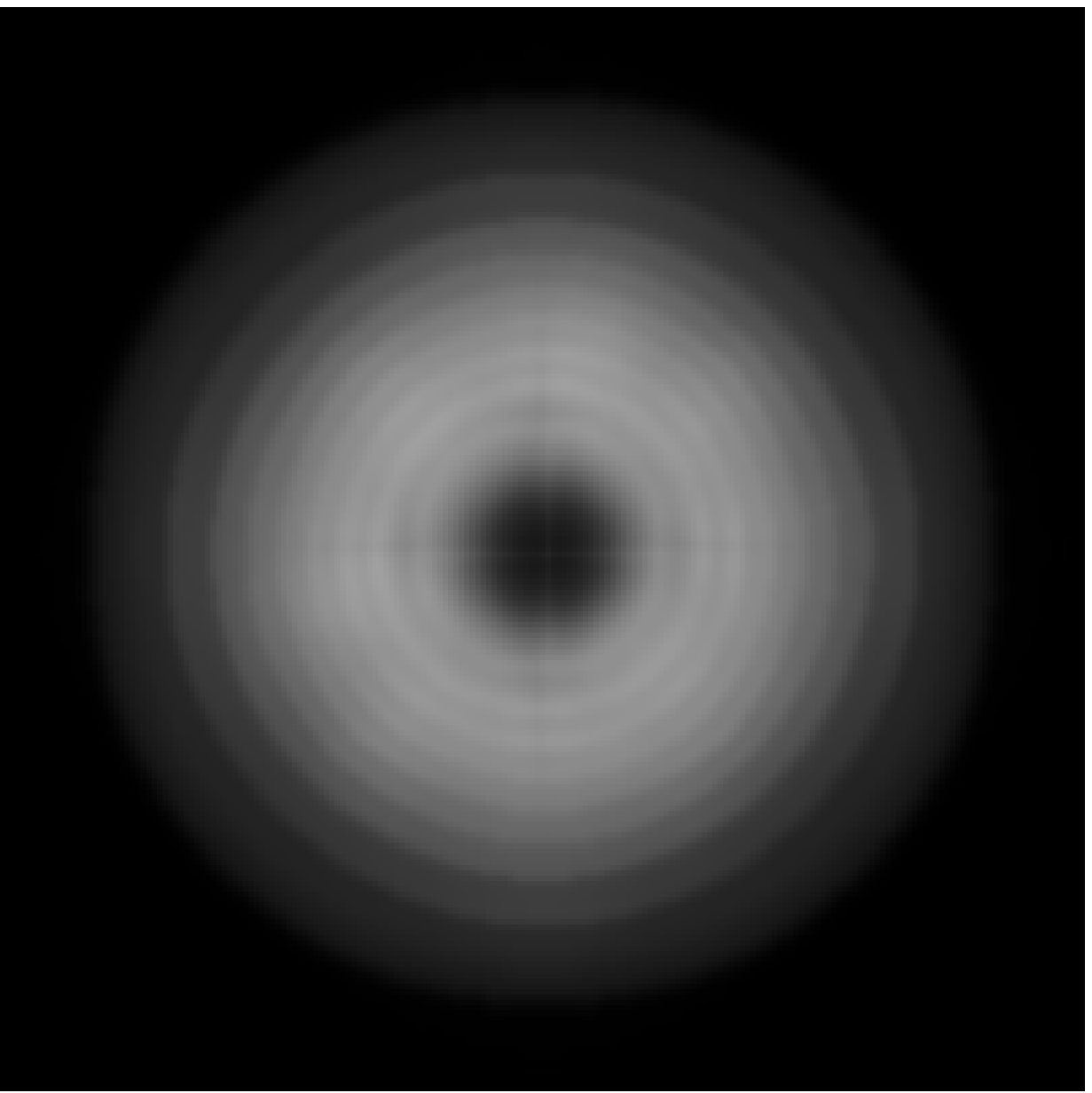} 
\includegraphics[width=0.15\textwidth]{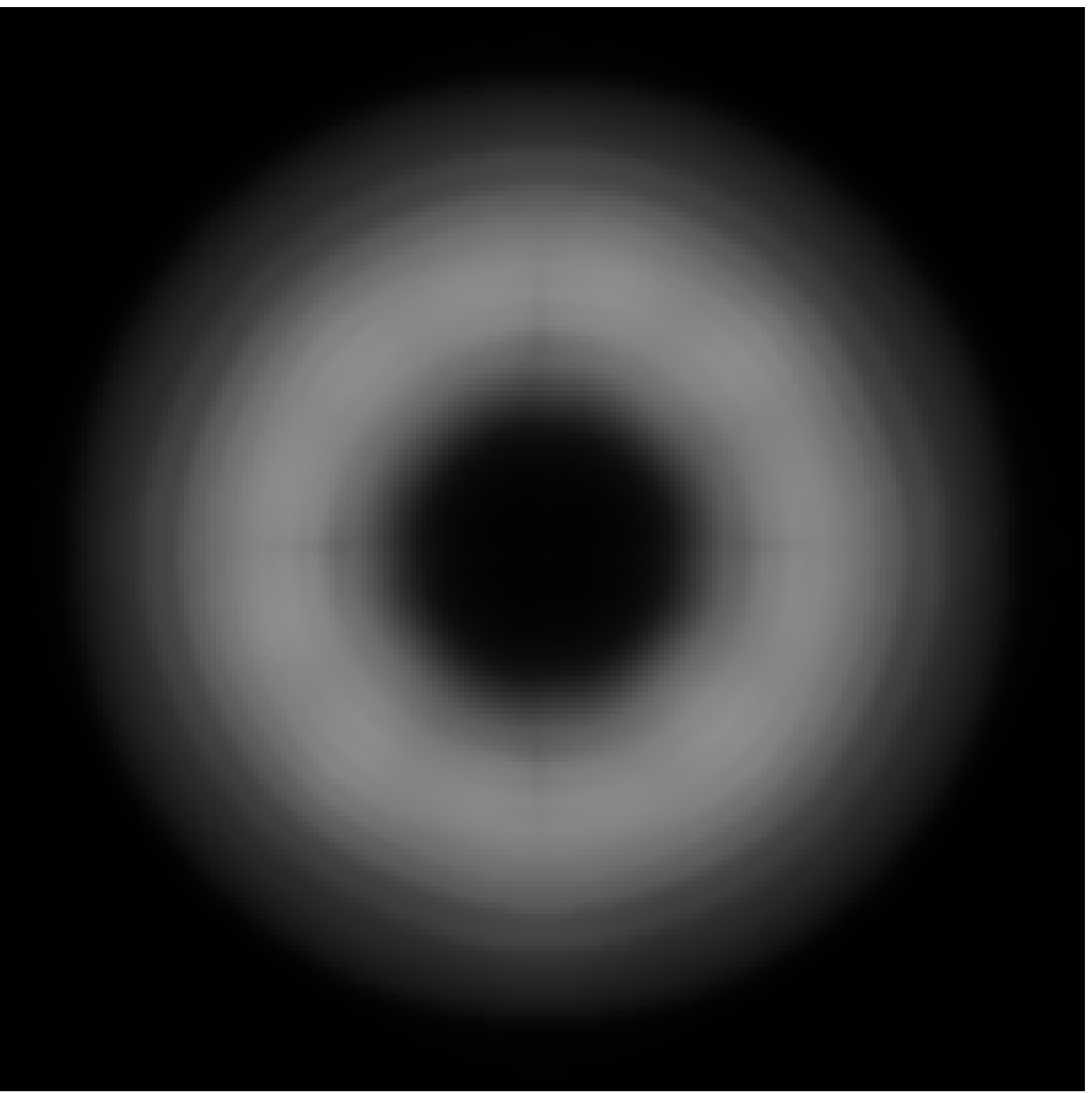} 
\includegraphics[width=0.15\textwidth]{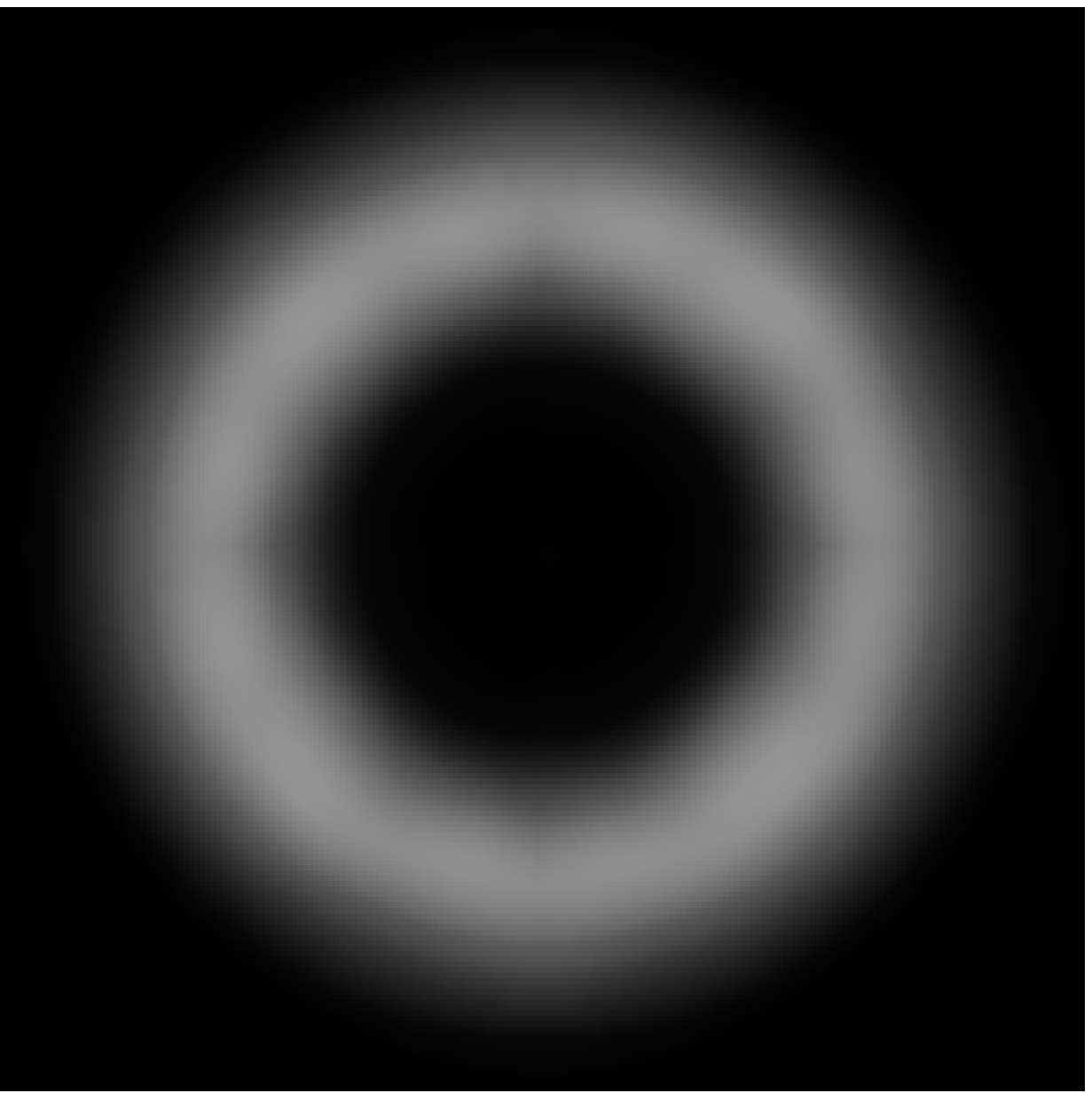} 
\includegraphics[width=0.15\textwidth]{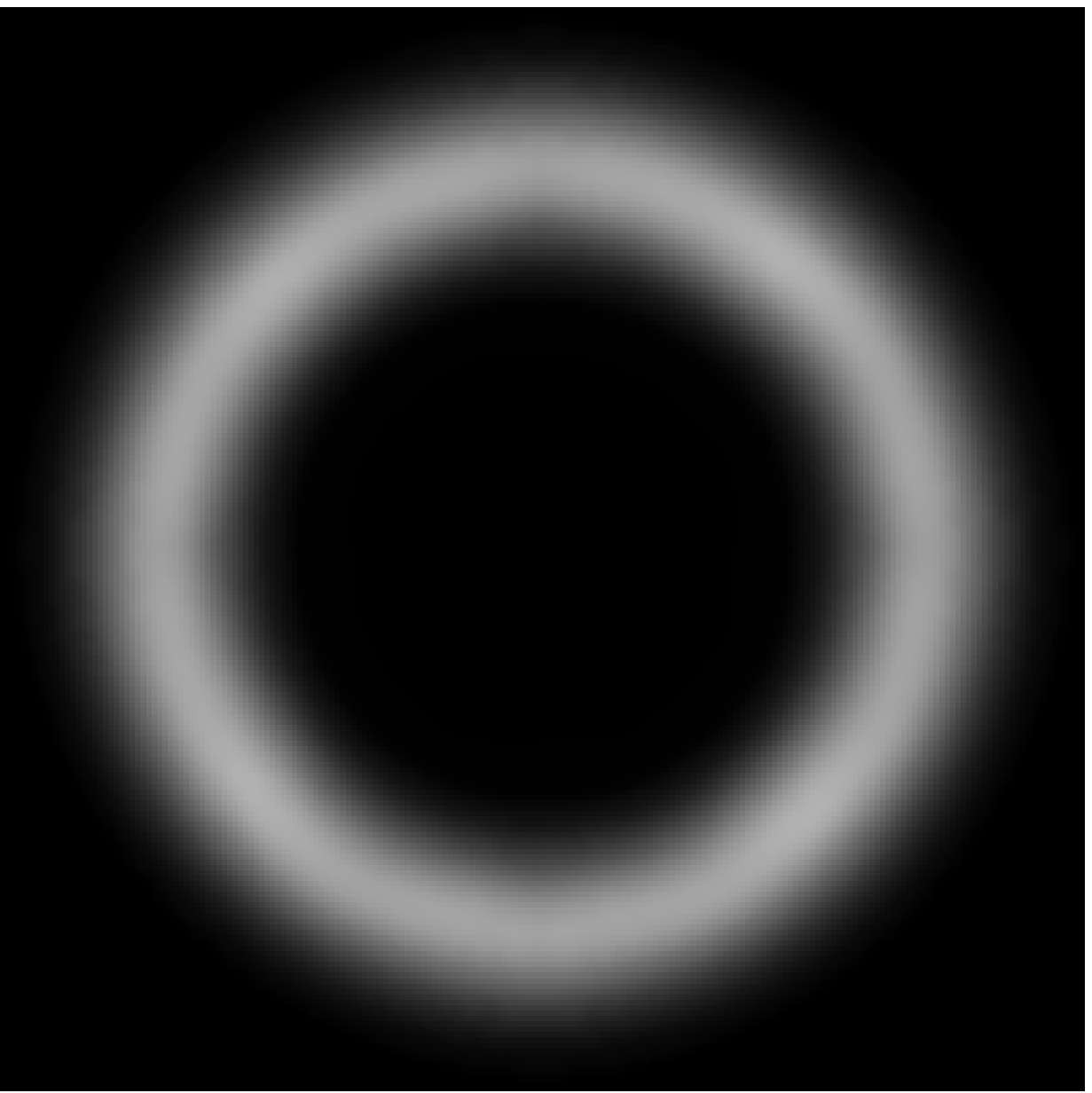} 
\includegraphics[width=0.15\textwidth]{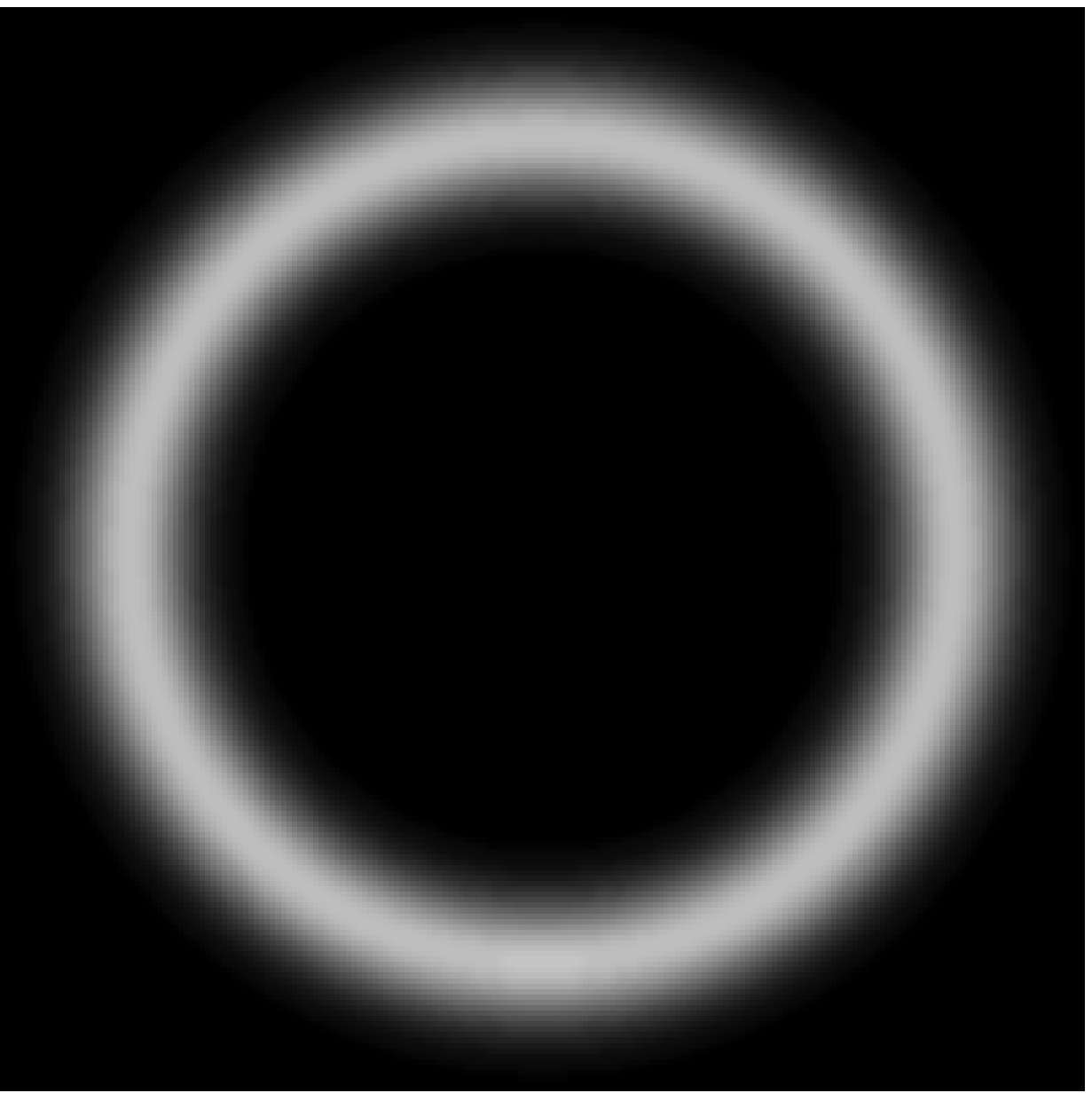} 
\caption{\label{fig:meta.dens} Minimizing metamorphoses between
densities with equal total mass. The optimal trajectories for $n_t$  are computed
between the first and last densities in each row. The remaining images
show $n_t$ at intermediate points in time.}
\end{figure}
\end{center}

We are here also in the semi-direct product case, the equations for
the conservation of
momentum being $L_\mg u + n\nabla z = \mathrm{cst} $ and $z
= z_0\circ g^{-1}$. Like for images, the constant in the first
conservation equation vanishes for horizontal geodesics in $G\circledS
N/G$. Optimal metamorphoses with densities are illustrated in Figure
\ref{fig:meta.dens}. 

\subsection{Plane Curves}
\label{sec:curves}
We here consider matching unit-length curves $\ga$ defined on the unit
circle 
$S^1$, represented, as in
\cite{you98,msj05,ymsm08} with their normalized tangent $\th\mapsto
\dot\ga_\th$ with $|\dot\ga_\th| = 1/2\pi$. The set $N$ is therefore a set
of functions $n : S^1 \to S^1(1/2\pi)$ where $S^1(r)$ is the
sphere with radius $r$ in $\mR^2$. We let $G$ be the group of
diffeomorphisms of $S^1$ and consider the reduced Lagrangian
$$
\ell(u, \nu) = \int_{S^1} \dot u_\th^2 d\th + \frac{1}{\sig^2}
\int_{S^1} |\nu|^2 d\th.
$$
We want to solve the metamorphosis problem while ensuring that curves
are closed, which
translates into:
$$
\int_S^1 n_t d\th = 0.
$$
To explicit \eqref{eq:meta.const}, we need a local chart to compute
the partial derivatives $\prt/\prt\nu$ and $\prt/\prt n$ ($N$ is not a
vector space here). Consider the representation $n = h_\al$ and $\nu =
\sig^2 \rho h_\al^\perp$ with $h_\al = (\cos\al, \sin\al)$ and $h^\perp_\al =
(-\sin\al, \cos\al)$. We then get the equations, with $\la_t\in\mR^2$,
\begin{equation}
\label{eq:meta.curv}
\left\{
\begin{array}{l}
\displaystyle 
-\frac{\prt^2 u_t}{\prt \th^2} + \rho_t \frac{\partial \al_t}{\prt \th} = 0\\ \\
\displaystyle 
\frac{\partial \rho_t}{\partial t} + \frac{\prt}{\prt\th}(u_t \rho_t) =
- \la_t^T h_{\al_t}^\perp \\
\\
\displaystyle
\dot \al_t = \sig^2 \rho_t - u_t\frac{\partial \al_t}{\prt \th}\\ \\
\displaystyle
\int_{S^1} \dot \al_t h_{\al_t}^\perp d\th = 0
\end{array}
\right.
\end{equation}

\begin{center}
\begin{figure}
\includegraphics[width=0.85\textwidth]{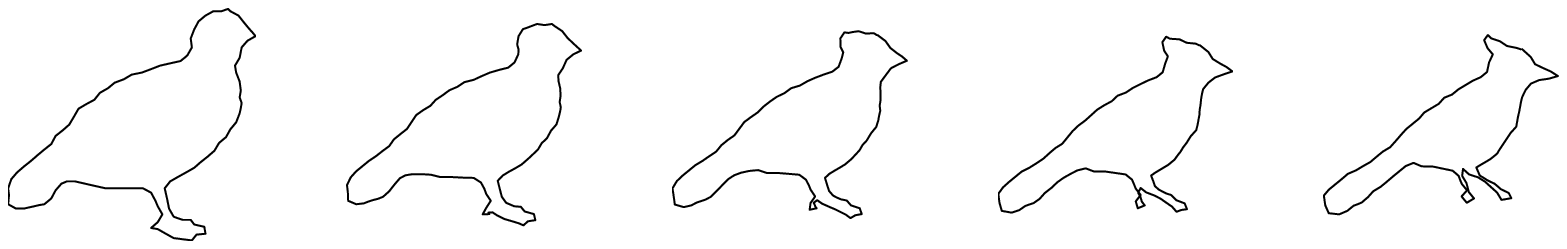} 
\includegraphics[width=0.85\textwidth]{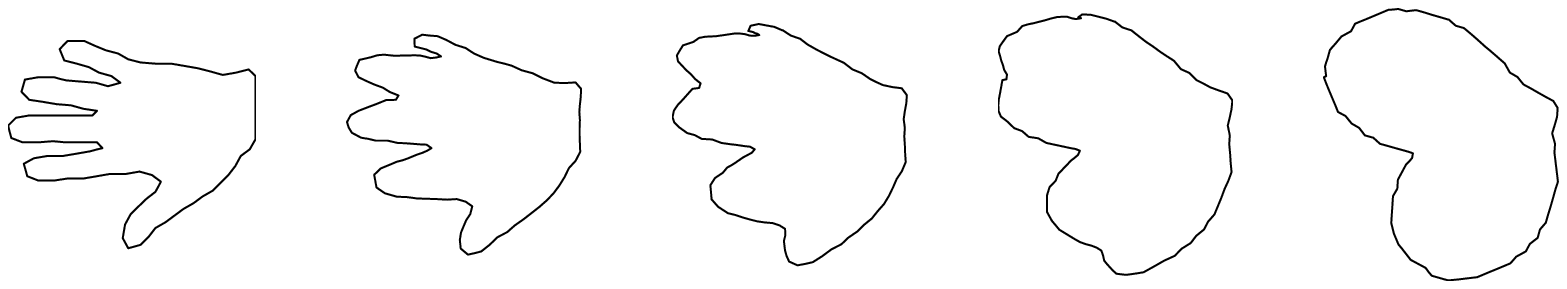} 
\caption{\label{fig:curv} Optimal metamorphoses between plane curves.The optimal trajectories for $n_t$  are computed
between the first and last curves in each row. The remaining images
show $n_t$ at intermediate points in time.}
\end{figure}
\end{center}

Interestingly, these equations can be notably simplified in the case
$\sig^2 = 1$, which has been considered in studied in
\cite{you98,ty00,ty01,ymsm08}. In this case, the change of variables $z_t^2 = \dot
g_t n_t\circ g_t$, where both sides are interpreted as complex numbers
reduces \eqref{eq:meta.curv} to the geodesic equations on a Grassmann
manifold, on which explicit computations can be made \cite{ymsm08}. The case $\sig^2
= 4$ is also interesting, and has been discussed in
\cite{jksj07}. Figure \ref{fig:curv} provides the result of curve
metamorphosis using $\sig^2=1$.

\section{Measure Metamorphoses}
\label{sec:meas}
We now focus on extending the example in section \ref{sec:dens} to include also singular
measures. The $L^2$ norm that we have used between densities is
therefore no longer available. We will rely on a construction that was
introduced in \cite{gty04}.

We let $H$ be a reproducing kernel Hilbert space
(RKHS) of
functions over $\mR^d$ and $N = H^*$; $H$ being an RKHS is equivalent to the fact
that for all $x\in \mR^d$, there exists a function $K_x \in H$ such
that, for all $f\in H$, $\scp{K_x}{f}_H = f(x)$. The kernel $K_H(x,y)
:= K_x(y)$ satisfies the equation $\scp{K_x}{K_y}_H = K_H(x,y)$. It also
provides an isometry between $N$ and $H$ via the relation $\eta
\mapsto K_H\eta$ with $\scp{K_H\eta}{f}_H = \lform{\eta}{f}$. This implies
in particular that the dual inner product on $N$ is given by
$$
\scp{\eta}{\tilde \eta}_N = \lform{\eta}{K_H\tilde\eta}.
$$

Letting $G$ be like in the previous section, we want to study
metamorphoses in $G\times N$. We define the action of $G$ on $N$ by
(with $g\in G, \eta\in N$ and $f\in H$)
$$\lform{g\eta}{f} = \lform{\eta}{f\circ g}.$$
This obviously generalizes the action on densities discussed in the
previous section.
 For $\eta\in N$ and $w\in\mathfrak g$, we have, for
all $f\in H$,
$\lform{w\eta}{f} = \lform{\eta}{\nabla f^Tw}$.

Since the action is linear, we can use the semi-direct product model
and the Lagrangian
$$
\ell(u, \nu) = \frac{1}{2}\|u\|_{\mathfrak g}^2 + \frac{1}{2\sig^2} \|\nu\|_N^2.
$$

To explicitly compute again \eqref{eq:meta.1} in this context, we need to
compute $(\partial \ell/\partial \nu)\diamond n$. We let $f = \de
\ell/\de \nu = (1/\sig^2) K_H\nu$. By definition, we have, for all
$w\in\mathfrak g$
\begin{eqnarray*}
\lform{f\diamond n}{w} &=& -\lform{f}{wn} \\
&=& -\lform{n}{\nabla f^Tw}\\
&=& -\lform{\nabla f \otimes n}{w}
\end{eqnarray*} 
We therefore obtain our first equation $(\de \ell/\de u) = \nabla f
\otimes \nu$. Since $\lform{f}{u\eta} = \lform{\eta}{\nabla f^T u}$,
the second equation, $\dot f_t + u_t \star f_t = 0$ is the advection:
$\dot f_t + \nabla^Tf_tu_t = 0$. This yields the system (with $L_H = K_H^{-1}$)
\begin{equation}
\label{eq:meta.meas1}
\left\{
\begin{array}{l}
\displaystyle L_\mg u_t =  \nabla f_t \otimes n_t \\
\\
\displaystyle \dot f_t + \nabla f_t^T u_t = 0\\ \\
\displaystyle
\dot n_t - u_tn_t = \sig^2 L_H f_t
\end{array}
\right.
\end{equation}

From the second equation, we get $f_t = f_0 \circ
g_t^{-1}$. Using $g\dot\eta = \nu = \sig^2 L_H f$ and $g\eta = n$,
we get
$$
n_t = g_t n_0 + \sig^2 g_t \int_0^t g_s^{-1} K_H^{-1}(f_0\circ g_s^{-1})ds.
$$
We therefore obtain integrated equations for measure metamorphoses
\begin{equation}
\label{eq:meta.meas}
\left\{
\begin{array}{l}
\displaystyle L_\mg u_t = \nabla f_t\otimes n_t \\ \\
\displaystyle
n_t = g_tn_0 + \sig^2 g_t \int_0^t g_s^{-1} L_H(f_0\circ g_s^{-1})ds
\end{array}
\right.
\end{equation}
with $\dot g_t = u_t\circ g_t$. 

We now pass to the theoretical study of the existence of solutions for
the initial value problem (IVP) and boundary value problem (BVP) for
measure metamorphosis (with uniqueness in the IVP case). The next two
sections are notably more technical than the rest of this paper. They
are well isolated from it, however, and it is possible, if
desired, to skip directly to section \ref{sec:ansatz}.

\section{Existence of solutions for the measure metamorphosis IVP}

\subsection{Hypotheses on the Hilbert spaces}
For the existence proofs to proceed, we need some conditions on the
Hilbert spaces $\mg$ and $H$. They are essentially adapted to $H$
being equivalent to $H^q_0$ (the completion, in $H^q$, of $C^\infty$ functions
with  compact support), in which case $H^- = H_0^{q-1}$ and $H^+ =
H_0^{q+1}$ ($q$ being large enough to ensure that $H^q$ is embedded in
$C^0$).

In the following, $\cst$ represent a generic constant, and $C$ a
generic continuous function of its parameters. 
We assume the existence of two spaces
$H^+$ and $H^-$ and the following properties, valid for some $q\geq 1$.

\noindent (H1) 
If $\tilde H = H^-, H$ or $H^+$, we have: if $f\in
\tilde H$ and $g \in C^q(\Om)$, then
$f\circ g\in \tilde H$ and
$$
\|f\circ g\|_{\tilde H} \leq C(\|g\|_{q, \infty}) \|f\|_{\tilde H}.
$$

\noindent (H2) 
For $f\in H^+$, and $g, \tilde g$ two $C^q$
diffeomorphisms 
$$\|f\circ g - f\circ \tilde g\|_{H} \leq\cst \|f\|_{H^+} 
C(\max(\|g\|_{q, \infty}, \|\tilde g\|_{q, \infty})) \|g-\tilde g\|_{q, \infty}.$$

\noindent (H3) For $f\in H$, define the operator $Q_f$ on $\mathfrak
g$ by $Q_fw = \nabla f^T w$. Then for all $f\in H$ and $g \in
C^{q}(\Om)$, $Q_f$ maps $\CX^q(\Om)$ to $H^-$ with 
$$\|Q_fw\|_{H^-} \leq \cst \|f\|_{H} \|w\|_{q, \infty}$$
 for all $w\in
\CX^q(\Om)$.
(Here, $\CX^q(\Om)$ denotes the set of $C^q$ vector fields on $\Om$
with the supremum norm of all derivatives of order less than $q$.)

\noindent (H4) If $f\in H^+$, then $K_H^{-1} f  \in (H^-)^*$ and for
all $z\in H^-$, $\lform{K_H^{-1} f}{z} \leq\cst
\|f\|_{H^+}\|z\|_{H^-}$.

\noindent (H5) $\mathfrak g$ is continuously included in $\CX^p_0(\Om)$ for
$p> q+1$, where $\CX^p_0$ is the completion of compactly supported
vector fields in $\CX^p(\Om)$.

Denote $N^+ = (H^-)^*$. We then have
\begin{theorem}
Under hypotheses (H1) to (H3), for all $T>0$, there exists a unique solution to
system \eqref{eq:meta.meas} over $[0,T]$ with initial conditions
$n_0\in N^+$ and $f_0\in H^{+}$.
\end{theorem}
\begin{proof}
We prove existence for small enough $T$ with a fixed point
argument. Consider the Hilbert space $L^2([0,T], \mg)$, with norm
$$
\|u\|_{2,T}^2 = \int_0^T \|u_t\|_\mg^2 dt.
$$
For $u\in
L^2([0,T], \mg)$, define $\Psi(u):=u'$ given by
\begin{equation}
\label{eq:meta.meas.2}
\left\{
\begin{array}{l}
\displaystyle u'_t = K_{\mg}(\nabla f_t\otimes n_t) \\ \\
\displaystyle
n_t = g_tn_0 + \sig^2 g_t \int_0^t g_s^{-1} L_H(f_0\circ g_s^{-1})ds
\end{array}
\right.
\end{equation}
with $\dot g_t = u_t\circ g_t$.

First note that the hypotheses imply that $\Psi$ is well
defined and takes values in $L^2([0,T], \mg)$. Indeed, by definition, 
\begin{eqnarray*}
\scp{u'_t}{w}_\mg &=&
\lform{\nabla f_t\otimes n_t}{w} \\
&=& \lform{n_t}{\nabla f_t^T w} \\
&\leq & \cst \|n_t\|_{N} \|f_t\|_{H^+} \|w\|_{\tilde q, \infty}
\\
&\leq & \cst \|n_t\|_{N} \|f_t\|_{H^+} \|w\|_{\mg}
\end{eqnarray*}
so that a sufficient condition for $\nabla f_t\otimes n_t \in \mg^*$
is $n_t\in N$ and $f_t\in H^+$. Since $f_t  = f_0 \circ
g_t^{-1}$, we have $\|f_t\|_{H^+} \leq \cst \|f_0\|_{H^+}
C(\|g_t^{-1}\|_{q, \infty})$. From \cite{ty05}, we have 
$\|g_t^{-1}\|_{q, \infty} = O(\|u\|_{2,T})$, yielding
\begin{equation}
\label{eq:bound.1}
\|f_t\|_{H^+} \leq \cst \|f_0\|_{H^+} C(\|u\|_{2,T}).
\end{equation}

For $z\in H$, we have
\begin{eqnarray*}
\lform{n_t}{z} &=& \lform{n_0}{z\circ g_t} + \sig^2 \int_0^t
\lform{K_H^{-1}f_s}{z_s \circ g_t \circ g_s^{-1}}ds
\\
&\leq& \left(\|n_0\|_{N} C(\|g_t\|_{q, \infty}) +  \sig^2
\int_0^t \|f_s\|_{H} C(\|g_t \circ g_s^{-1}\|_{q, \infty}) ds\right)\|z\|_{H}
\end{eqnarray*}
so that 
\begin{equation}
\label{eq:bound.2}
\|n_t\|_{N} \leq \left(\|n_0\|_{N} +  \sig^2
t \|f_0\|_{H}\right) C(\|u\|_{2,T}).
\end{equation}
This implies that
$$
\|u'_t\|_{\mg} \leq  \|f_0\|_{H^+} \left(\|n_0\|_{N} +  \sig^2
t \|f_0\|_{H} \right) C(\|u\|_{2,T})
$$
and
$$
\|u'\|_{2,T} \leq \cst \sqrt{T} \|f_0\|_{H^+} \left(\|n_0\|_{N} +  \sig^2
T \|f_0\|_{H} \right) C(\|u\|_{2,T}).
$$
In particular, this implies that, for any $M>0$, there exists a $T_0(M)$
(only depending on $\|n_0\|_{N^+}$ and $\|f_0\|_{H^+}$)
such that, for $T<T_0$,  $\|u\|_{2,T}\leq M$ implies $\|u'\|_{2,T}
\leq M$. From now on, we
assume that $T<T_0(M)$ with $M=1$.

Note that a similar computation also shows that $n_t \in N^+$: for
$z\in H^-$, we have 
\begin{eqnarray*}
\lform{n_t}{z} &=& \lform{n_0}{z\circ g_t} + \sig^2 \int_0^t
\lform{K_H^{-1}f_s}{z_s \circ g_t \circ g_s^{-1}}ds
\\
&\leq& \left(\|n_0\|_{N^+} C(\|g_t\|_{q, \infty}) +  \sig^2
\int_0^t \|f_s\|_{H^+} C(\|g_t \circ g_s^{-1}\|_{q, \infty}) ds\right)\|z\|_{H^-}
\end{eqnarray*}
so that 
\begin{equation}
\label{eq:bound.2.2}
\|n_t\|_{N^+} \leq (\|n_0\|_{N^+} +  \sig^2
t \|f_0\|_{H^+} )C(\|u\|_{2,T}).
\end{equation}

We now ensure that $\Psi$ is contractive. Take $u, \tilde u$ with
$\max(\|u\|_{2,T}, \|\tilde u\|_{2,T})\leq 1$. We want to show that
$T$ can be chosen so that $\|u' - \tilde u'\|_{2,T} \leq \rho \|u -
\tilde u\|_{2,T}$ with $\rho < 1$, where $u' = \Psi(u)$ and $\tilde u'
 = \Psi(\tilde u)$. We have
\begin{eqnarray*}
\|u'_t - \tilde u'_t\|_{\mathfrak g} &=& \|\nabla f_t \otimes n_t -
\nabla \tilde f_t \otimes \tilde n_t\|_{\mathfrak g^*} \\
&\leq& \|\nabla (f_t - \tilde f_t) \otimes n_t\|_{\mathfrak g^*} +
\|\nabla \tilde f_t \otimes (n_t - \tilde n_t)\|_{\mathfrak g^*} \\
&\leq& \cst \|n_t\|_{N^+} \|f_t - \tilde f_t\|_{H} + \cst \|n_t -
\tilde n_t\|_{N} \|\tilde f_t\|_{H^+}
\end{eqnarray*}
(Here, we have used the fact that for $n\in N^+$ and $z\in H$, we have both
$\lform{n}{z}\leq \|n\|_N\|z\|_H$ and $\lform{n}{z} \leq \|n\|_{N^+} \|z\|_{H^-}$.)

Upper bounds for $\|\tilde f_t\|_{H^+}$, $\|n_t\|_N$ and 
$\|n_t\|_{N^+}$ are provided by equations \eqref{eq:bound.1},
\eqref{eq:bound.2} and \eqref{eq:bound.2.2}.

Moreover, from (H2), we have
$$\|f_t - \tilde f_t\|_{H} \leq C(\max(\|g^{-1}_t\|_{q, \infty},
\|\tilde g^{-1}_t\|_{q, \infty}) \|f_0\|_{H^{+}}
\|g_t^{-1} - \tilde g_t^{-1}\|_{q, \infty}$$
 and, using \cite{ty05},
we have $\|g_t^{-1} - \tilde g_t^{-1}\|_{q+1, \infty} \leq \cst  \|u-\tilde u\|_{2,T}$ so that
\begin{equation}
\label{eq:bound.3}
\|n_t\|_{N^+} \|f_t - \tilde f_t\|_{H} \leq \cst (\|n_0\|_{N^+} +  \sig^2
t \|f_0\|_{H^+}) \|f_0\|_{H^{+}} \|u-\tilde u\|_{2,T}
\end{equation}

For $z\in H$, we can write
\begin{eqnarray*}
\lform{n_t-\tilde n_t}{z} &=& \lform{n_0}{z\circ g_t - z\circ \tilde
g_t} \\
&& + \sig^2 \int_0^t \lform{K_H^{-1} f_s}{z\circ g_t \circ g_s^{-1} -
z\circ \tilde g_t\circ \tilde g_s^{-1}} ds \\
 && + \sig^2 \int_0^t \lform{K_H^{-1} (f_s-\tilde f_s)}{z\circ \tilde g_t\circ \tilde g_s^{-1}} ds \\
\text{(i)} &\leq&  \cst\,\|n_0\|_{N^+} \|z\|_H\|g_t - \tilde g_t\|_{q, \infty} \\
\text{(ii)} && + \ \cst\, \sig^2 t \|f_0\|_{H^+} \|z\|_H \max_s\|g_t \circ g_s^{-1} -
\tilde g_t\circ \tilde g_s^{-1}\|_{q, \infty} \\
\text{(iii)} && + \ \cst\, \sig^2 t \|f_0\|_{H^+} \|z\|_H\max_s \|g_s -\tilde g_s\|_{q, \infty}
\|\tilde g_t\circ \tilde g_s^{-1}\|_{q, \infty}.
\end{eqnarray*}
For (i), we have used 
\begin{multline*}
\lform{n_0}{z\circ g_t - z\circ \tilde
g_t} \\
\leq \|n_0\|_{N^+} \|z\circ g_t - z\circ \tilde
g_t\|_{H^-} \leq \cst \|n_0\|_{N^+} \|z\|_H \|g_t - \tilde g_t\|_{p,
\infty}.
\end{multline*}
 For (ii), we have used the same argument combined with the fact
that, since $f_s\in H^+$, $K_H^{-1} f_s\in N^+$ with $\|K_H^{-1}
f_s\|_{N^+} \leq \cst \|f_s\|_{H^+}$. For (iii), the computation uses
the fact that for $\tilde z\in H$,
$$\lform{K_H^{-1} (f_s-\tilde f_s)}{\tilde z} \leq \|f_s-\tilde f_s\|_H
\|z\|_H.$$

We therefore obtain the inequality
\begin{equation}
\label{eq:bound.4}
\|n_t-\tilde n_t\|_H \leq \cst (\|n_0\|_{N^+} + \|f_0\|_{H^+}) \|u -
\tilde u\|_{2,T}.
\end{equation}

Collecting the previous estimates, we have
$$
\|u'_t - \tilde u'_t\|_\mg \leq F(\|n_0\|_{N^+}, \|f_0\|_{H^+}) \|u -
\tilde u\|_{2,T},
$$
where $F$ is a polynomial. This implies 
$$
\|u' - \tilde u'\|_{2,T} \leq \sqrt{T} F(\|n_0\|_{N^+}, \|f_0\|_{H^+}) \|u -
\tilde u\|_{2,T},
$$
so that $\Psi$ is contractive for small enough $T$.

The extension from small times to all times can be done like in
\cite{ty05}, and we only sketch the details. According to the
small time computation, the length over which the solution exists is
at least the inverse of a polynomial function of $\|n_0\|_{N^+}$ and
$\|f_0\|_{H^+}$. One will therefore be able to extend this equation
beyond $T$, unless either $\|n_t\|_{N^+}$ or $\|f_t\|_{H^+}$ tends to
infinity when $t$ tends to $T$. From \eqref{eq:bound.1} and
\eqref{eq:bound.2}, this can happen only if $\|u\|_{2,t}$ tends to
infinity when $t$ tends to $T$. But this is impossible, because
\eqref{eq:meta.meas} is a geodesic equation on $G\circledS N$ which
implies that the value of $h_t := \|u_t\|_\mg^2 + (1/\sig^2)
\|\nu_t\|^2_{N}$ is constant over time. This implies in particular
that $\|u\|_{2,t}^2 \leq t h_0$ and therefore cannot tend to infinity
in finite time.
\end{proof}

\section{Existence of solutions for the measure metamorphosis BVP}

Our goal in this section is to prove that, under some conditions on
$H$, the boundary value problem for measure metamorphoses (BVP) always
has solutions. This problem requires to minimize, with fixed $n_0$ and
$n_1$,
$$
E(u,n) := \int_0^1 \|u_t\|_{\mg}^2 dt + \frac{1}{\sigma^2} \int_0^1
\|\dot n_t - u_tn_t\|_N^2 dt.
$$

Letting $f_t = K_H(\dot n_t - u_tn_t)$, the problem is equivalent to minimize
$$
E(u,f) := \int_0^1 \|u_t\|_{\mg}^2 dt + \sigma^2 \int_0^1 \|f_t\|_H^2 dt
$$
with boundary  
$$
n_1 = g_1n_0 + \sig^2 g_1 \int_0^1 g_s^{-1} (K_H^{-1}f_s) ds.
$$
We have
\begin{theorem}
Assume that (H1) and (H2) hold and that $H^+$ is dense in $H$. Then,
for any given $n_0, n_1\in H$, there exists a minimizer for the BVP.
\end{theorem}
\begin{proof}
Consider a
minimizing sequence $(u^{(k)}, f^{(k)})$. Using a time change if
necessary, we can ensure that $c_k := \|u_t^{(k)}\|_\mg^2 + \sig^2
\|f_t^{(k)}\|^2$ is independent of time and therefore bounded. Moreover, we can extract a subsequence (still denoted 
$(u^{(k)}, f^{(k)})$) which weakly converges to some $(u,f)$ in
$L^2([0,1], \mg\times H)$, equipped with the norm $\|(u',f')\| =
E(u',f')$. This implies that $E(u,f)  \leq \liminf E(u^{(k)},
f^{(k)})$ so that the only thing that needs to be showed is that the
boundary condition is still satisfied, namely, letting 
$$
n'_1 = g_1n_0 + \sig^2 g_1 \int_0^1 g_s^{-1} (K_H^{-1}f_s) ds
$$
with $\dot g_t = u_t\circ g_t$, we have $n'_1 = n_1$. For this, it
suffices to show that, for $z$ in a dense subset of $H$, we have
$\lform{n'_1}{z} = \lform{n_1}{z}$. Since, for all $k$, we have
$$
n_1 = g^{(k)}_1n_0 + \sig^2 g^{(k)}_1 \int_0^1 (g^{(k)}_s)^{-1} (K_H^{-1}f^{(k)}_s) ds,
$$
it suffices to show that, for $z$ in a dense subset of $H$, we have
\begin{equation}
\label{eq:1}
\lform{n_0}{z\circ g^{(k)}_1} \to \lform{n_0}{z\circ g_1}
\end{equation}
and
\begin{equation}
\label{eq:2}
\int_0^1 \lform{K_H^{-1}f^{(k)}_s}{z\circ g^{(k)}_1\circ
(g^{(k)}_s)^{-1}} ds \to \int_0^1 \lform{K_H^{-1}f_s}{z\circ
g_1\circ g_s^{-1}}  ds 
\end{equation}

Because of the weak convergence of $u^{(n)}$ to $u$, the flows
$g^{(n)}$ converge to $g$ for the $ (p-1, \infty)$-norm, uniformly in
time \cite{ty05}. Taking $z\in H^+$, we have 
\begin{multline*}
\scp{n_0}{z\circ g^{(k)}_1 -  z\circ g_1} \leq\\
C(\sup_t(\|g_t^{(k)}\|_{p, \infty},\|g_t\|_{p, \infty}))  \|n_0\|_{H^*}\|z\|_{H^+} \sup_t \|g^{(k)}_t - g_t\|_{p, \infty}
\end{multline*}
which tends to 0 so that  
\eqref{eq:1} is true for all $z\in H^+$.  Splitting the terms in \eqref{eq:2},
it suffices to show that for all $s\in [0,1]$, 
\begin{equation}
\label{eq:2.1}
\int_0^1 \lform{K_H^{-1}f^{(k)}_s}{z\circ g^{(k)}_1\circ
(g^{(k)}_s)^{-1}}  ds - \int_0^1 \lform{K_H^{-1}f^{(k)}_s}{z\circ
g_1\circ g_s^{-1}}  \to 0 
\end{equation}
and
\begin{equation}
\label{eq:2.2}
\int_0^1 \lform{K_H^{-1}f^{(k)}_s}{z\circ g_1\circ
(g_s)^{-1}}ds  - \int_0^1 \lform{K_H^{-1}f_s}{z\circ
g_1\circ g_s^{-1}}ds  \to 0. 
\end{equation}
For \eqref{eq:2.1}, and because
we have ensured that $\|f^{(k)}_s\|_H$ and $\|u^{(k)}_s\|_\mg$ are uniformly
bounded, it suffices to show that each term in the integral tends to 0
and then use the dominated convergence theorem.
The left-hand term of \eqref{eq:2.1} is bounded in absolute value by
$$\cst \|f^{(k)}_s\|_H\|z\|_{H^+} \|g^{(k)}_1\circ
(g^{(k)}_s)^{-1} - g_1\circ
g_s^{-1}\|_{q, \infty}$$
which tends to 0, so that \eqref{eq:2.1}
holds. For \eqref{eq:2.2} we only need the fact that $z\circ
g_1\circ g_s^{-1}$ belongs to $H$ and the weak convergence of
$f^{(k)}$ to $f$ to conclude.

This shows that $n'_1=n_1$ and concludes the proof of the theorem.
\end{proof}

\subsection{Remark}

Equations \eqref{eq:meta.meas} have been obtained from general
formulae that were derived under the assumption that $G$ is a Lie group,
(which is not the case here). It is important to rigorously recompute
the Euler equation to reconnect the IVP and the BVP. The variation
with respect to $u$ is straightforward and provides the first equation
in \eqref{eq:meta.meas}.

We now discuss the
 minimization in $n$ with
fixed $u$. Letting $\eta = g^{-1}n$ the problem with fixed $u$ (and
therefore fixed $g$) is equivalent to the minimization of
$$
F(\eta) = \int_0^1 \|g_t \dot\eta_t\|_N^2 dt
$$
with fixed boundary conditions $\eta_0=n_0$ and $\eta_1 = g_{1}^{-1}
n_1$. 

Assume the following hypothesis:\\

\noindent (H1b) $z\to z\circ g$ is weakly continuous for $z\in H$
(where $H^*=N$)  \\

For $g\in G$, introduce the operator $K^g_H$ defined by
$K_H^g\eta =K_H(g\eta)\circ g$. With this notation, 
$$
F(\eta) = \int_0^1 \lform{\dot\eta_t}{K_H^{g_t} \dot\eta_t}_N dt.
$$
Let $L_H^g = (K_H^g)^{-1}$, i.e., $L_H^g f = g^{-1}.L_H(f\circ g^{-1})$, and denote
\begin{equation}
\label{eq:bar.l}
\bar L_H^g = \int_0^1 L_H^{g_t} dt.
\end{equation}

We have for any $f\in H$,
\begin{equation}
  \label{eq:alain1}
  \lform{\bar L_H^g f}{f}=\int_0^1 
  \lform{L_H^{g_t}f}{f}dt =\int_0^1 \|f\circ g_t^{-1}\|_H^2 dt\geq C 
\|f\|^2_H
\end{equation}
where the last inequality comes from (H1) with 
$C=[\sup_{t}C(\|g_t^{-1}\|_{q,\infty})]^{-1}$.
Thus, $\bar L_H^g f=0$ if and only if $f=0$
 and $\bar L_H^g(H)$ is dense in $H^*$. 
If we prove that $\bar
 L_H^g(H)$ is closed, we will get that $\bar
 L_H^g(H):H\to H^*$ is invertible. Let $(f_n)_{n\geq 0}$ be a sequence
 in $H$ and $\eta\in H^*$ such that $\bar L_H^g(f_n)\to \eta$ in
 $H^*$. Then we get from \eqref{eq:alain1} that $f_n$ is bounded in $H$ and
 we can assume that it admits a weak limit $f_\infty$ in $H$. Thus
 $\lform{\bar L_H^g f_\infty}{f}=\int_0^1 \langle f_\infty\circ
 g_t^{-1}, f\circ g_t^{-1}\rangle_H dt =\lim\int_0^1 \langle f_n\circ
 g_t^{-1}, f\circ g_t^{-1}\rangle_H dt$  where the last equality
 comes from (H1b) and the dominated convergence theorem. This yields
$\lform{\bar L_H^g f_\infty}{f}=\lform{\eta}{f}$ for any $f\in H$ so
that $\eta=\bar L_H^g f_\infty$ and $\bar L_H^g(H)$ is closed.

Let $f_0$ such that $\bar  L_H^gf_0=\eta_1-\eta_0$ and define
\begin{equation}
\tilde{\eta}_t=\eta_0+\int_0^tL_H^{g_s}f_0ds\label{eq:min.eta}
\end{equation}
for any $t\in [0,1]$. If we can prove
that for any $(\dot{\eta}_t)\in L^2([0,1],H^*)$ with $\eta_1$ and
$\eta_0$ fixed
$$  F(\eta-\tilde{\eta})=F(\eta)+\cst
$$
we deduce immediately the result. However, we have 
$$
F(\eta-\tilde{\eta})=F(\eta)+F(\tilde{\eta})-2\int_0^1\lform{\dot{\eta}_t}{K_H^{g_t}\dot{\tilde{\eta}}_t}dt$$
and
$\int_0^1\lform{\dot{\eta}_t}{K_H^{g_t}\dot{\tilde{\eta}}_t}dt=\int_0^1\lform{\dot{\eta}_t}{f_0}dt=\lform{\eta_1-\eta_0}{f_0}$
so that the result is proved.

We therefore have proved that minimizing $E$ with respect to $u,n$ is
the same as minimizing
$$
\tilde E(u) = \int_0^1 \|u_t\|^2_\mg dt + \frac{1}{\sig^2} \lform{\eta_1-\eta_0}{(\bar L^g_H)^{-1}
(\eta_1-\eta_0)}
$$
with respect to $u$.
We also have the expression \eqref{eq:min.eta} for the optimal $\eta$
which is consistent with the second equation in \eqref{eq:meta.meas}. 

\section{A computational ansatz for point measure matching}
\label{sec:ansatz}
The previous theorems provide a rigorous foundation for the considered
measure matching approach. However, an important issue 
needs to be addressed. The space $N$, which has been introduced in
order to take advantage of its Hilbert structure, is a very big space
that contains distributions that are more singular than measures. Now,
when matching two measures $n_0$ and $n_1$, the  question naturally
arises of whether the optimal evolution, i.e., the measure $n_t$, can
turn  up being more singular than measures, since the existence
theorem only ensures that it belongs to $N$. 

The second equation in \eqref{eq:meta.meas} indicates that this should
not be the case, since it says that
$$
n_t = g_t n_0 + g_t \int_0^t L_H^{g_s} (\bar L_H^g)^{-1} (g_1^{-1} n_1 -
n_0) ds
$$
where $L^g_H$ and $\bar L_H^g$ are defined above and in equation
\eqref{eq:bar.l}. 
Thinking of $L_H$ as a differential operator, the number of
derivatives computed by $L_H^{g_s}$ is ``canceled'' by the
pre-application of $(\bar L_H^g)^{-1}$ so that $n_t$ should not be
more singular than $(g_1^{-1} n_1 -
n_0)$ which is a measure. It is therefore reasonable to conjecture
that when $n_0$ and $n_1$ are weighted sums of Dirac measures (which
is a case of practical interest), then $n_t$ is a measure which has an
absolutely continuous part, and a singular part which is also a sum of
Diracs. More precisely, if
$$
n_0 = \sum_{k=1}^q \al_k^{(0)} \de_{x_k^{(0)}}, \qquad n_1 = \sum_{k=1}^r \be_k^{(1)}
\de_{y_k^{(1)}},
$$
a reasonable ansatz for $n_t$ is 
$$
n_t = \sum_{k=1}^r \al_k(t) \de_{x_k(t)} + \sum_{k=1}^r \be_k(t) \de_{y_k(t)} + f(t, .)dx.
$$

Assuming this, the Lagrangian $\ell(v_t, n_t, \nu_t)$ can be
considered as a function of $(\al_k(t))$, $(\be_k(t))$, $(x_k(t))$,
$(y_k(t))$, $f(t,.)$ and their time derivatives, with an explicit
expression in terms of the kernel $K_H$ and its space derivatives that
we do not provide here, since it is quite lengthy. Minimization can
then be done with standard methods, with  boundary conditions
$\al_k(0) = \al_k^{(0)}$, $\be_k(0) = 0$, $\al_k(1) = 0$, $\be_k(1) =
\be_k^{(1)}$, $x_k(0) = x_k^{(0)}$, $y_k(1) = y_k^{(1)}$ and $f(0,.) =
f(1,.) = 0$.

\section{More Metamorphosis}
Without getting into the level of rigor and detail developed with
measure metamorphosis, we now review additional situations in which
metamorphoses can be used. At the difference of the examples
considered in section \ref{sec:expl}, the following models have not
been solved numerically yet, nor has a theoretical analysis been
developed although we expect that the previous proofs of
existence of solutions can be modified to work in these cases also.

\subsection{Singular Image Metamorphosis}
In section \ref{sec:meas}, we have extended density metamorphosis to a context
that includes singular
measures. A similar analysis can be made to extend image metamorphosis
to generalized functions. Let $H$ be a space of smooth scalar
functions and $N = H^*$  as before. The extension to $N$ of the action
of diffeomorphisms on images is $(g,n) \mapsto gn$ with
$$
\lform{gn}{f}  = \lform{n}{\det(Dg) f\circ g},
$$
the infinitesimal action being $\lform{un}{f} =
\lform{n}{\mathrm{div}(fu)}$. Take as before the simplest reduced
Lagrangian
$$
\ell(u, \nu) = \|u\|_\mg^2 + \frac{1}{\sig^2} \|\nu\|_N^2.
$$

We have, letting $f = (1/\sig^2)K_H\nu$,
\begin{eqnarray*}
\lform{\frac{\de \ell}{\de \nu} \diamond n}{u} &=&  \lform{f\nabla
n}{u}\\
\text{ and } \lform{u\star \frac{\de \ell}{\de \nu}}{\om}  &=&
\lform{\om}{\mathrm{div}(fu)}
\end{eqnarray*}
with the notation $\lform{\nabla n}{w}  = -
\lform{n}{\mathrm{div}w}$. We therefore obtain the generalized version
of \eqref{eq:meta.img}:
\begin{equation}
\label{eq:meta.gen.im}
\left\{
\begin{array}{l}
\displaystyle L_\mg u_t =  - f_t  \nabla n_t \\
\\
\displaystyle \dot f_t + \mathrm{div} (f_t u_t) = 0\\ \\
\displaystyle
\dot n_t - u_tn_t = \sig^2 L_H f_t
\end{array}
\right.
\end{equation}

Like for measures, this leads to integrated equations
\begin{equation}
\label{eq:meta.im.gen.2}
\left\{
\begin{array}{l}
\displaystyle L_\mg u_t = - f_t \nabla n_t \\ \\
\displaystyle
n_t = g_tn_0 + \sig^2 g_t \int_0^t g_s^{-1} L_H\left(f_0\circ g_s^{-1} \det(g_s^{-1})\right)ds
\end{array}
\right.
\end{equation}
with $\dot g_t = u_t\circ g_t$.

In the 1D case, with $\rho = \sig f$ and
$m=L_{\mg}u=(1-\partial_x^2)u$, we get a new version of
\eqref{SDPsystem1D}:
\begin{eqnarray}
&&\hspace{5mm}
\partial_t m+u\partial_x m + 2m\partial_xu = - \rho\partial_x L_H\rho
\quad\hbox{with}\quad
\partial_t\rho + \partial_x(\rho u) = 0
\label{SDPsystem1D.gen}
\end{eqnarray}

\subsection{Smooth Image Metamorphosis}

We can go in the opposite direction and consider norms that will apply
to smooth images in the metamorphosis formulation. Namely, keeping the
notation of the previous section, we can define
$$
\ell(u, \nu) = \|u\|_\mg^2 + \frac{1}{\sig^2} \|\nu\|_H^2.
$$
Since the situation is completely symmetrical, we can immediately
write the new system, with $f = (1/\sig^2)L_H\nu$,
\begin{equation}
\label{eq:meta.smooth.im}
\left\{
\begin{array}{l}
\displaystyle L_\mg u_t =  - f_t  \nabla n_t \\
\\
\displaystyle \dot f_t + \mathrm{div} (f_t u_t) = 0\\ \\
\displaystyle
\dot n_t - u_tn_t = \sig^2 K_H f_t
\end{array}
\right.
\end{equation}
The second equation must be understood in a generalized sense, $f$
being advected by the flow as a measure.

An interesting feature of this system is that it admits singular
solutions for the pair $(L_\mg u, f)$. Indeed, assume that $f_0 =
\sum_{k=1}^Q w^{(k)} \de_{x^{(k)}_0}$ is a sum of weighted point masses. The
second equation in \eqref{eq:meta.smooth.im} implies that 
$$
f_t = \sum_{k=1}^Q w^{(k)} \de_{x^{(k)}_t}
$$
with $\dot x^{(k)}_t = u_t(x^{(k)}_t)$. We also have
$$
L_\mg u_t = \sum_{k=1}^Q a^{(k)}_t \otimes \de_{x_t^{(k)}}
$$
with $a_t^{(k)} = w^{(k)} \nabla n(x_t^{(k)})$ and $n$ evolves
according to the last equation in the system.

Exploiting this very simple structure can lead to interesting
new methods for the analysis of smooth (or smoothed) images and
will be considered in future work. The 1D evolution equation
associated to this context is (letting, again, $\rho=\sig f$)
\begin{eqnarray}
&&\hspace{5mm}
\partial_t m+u\partial_x m + 2m\partial_xu = - \rho\partial_x K_H\rho
\quad\hbox{with}\quad
\partial_t\rho + \partial_x(\rho u) = 0
\label{SDPsystem1D.smooth}
\end{eqnarray}

\subsection{Smooth Densities}
Obviously, the same reduced Lagrangian can be used with
densities. Using $\lform{\nabla f}{u} = - \lform{f}{\text{div}(u)}$ as
a definition of $\nabla f$ for generalized functions,
we get the system:
\begin{equation}
\label{eq:meta.meas.smooth}
\left\{
\begin{array}{l}
\displaystyle L_\mg u_t =  - \nabla n_t f_t -  n_t \nabla f_t \\
\\
\displaystyle \dot f_t + u_t^T \nabla f_t = 0\\ \\
\displaystyle
\dot n_t +\text{div}( u_tn_t) = \sig^2 K_H f_t
\end{array}
\right.
\end{equation}
Here also, singular solutions in $(L_\mg u, f)$ exist, although (as
seen from the first equation), $L_\mg u$ now is one derivative less
regular than $f$. 

\section{Conclusion}

We have provided a general framework for the pattern matching theory
of metamorphoses. The provided equations are quite versatile, and
adapted to any context in which a Lie group acts on a manifold.

In the particular case of diffeomorphisms acting on generalized functions, we have
obtained a new set of equations, and showed that they had solutions,
and so does the initial variational problem. Our equations seem to
indicate that, when matching measures, metamorphoses do not generate
additional singularities, but that they may introduce smooth components
that did not appear in the initial problem. 

Open for future work is the interesting problem of building an
efficient numerical implementation of measure metamorphoses, and their
use in specific pattern matching applications.

\bibliographystyle{plain}
\bibliography{defor}

\begin{thebibliography}{10}

\bibitem{arn66}
V.~I. Arnold.
\newblock Sur un principe variationnel pour les ecoulements stationnaires des
  liquides parfaits et ses applications aux probl\`emes de stanbilit\'e non
  lin\'eaires.
\newblock {\em J. M\'ecanique}, 5:29--43, 1966.

\bibitem{bb82}
R~Bajcsy and C~Broit.
\newblock Matching of deformed images.
\newblock In {\em The 6th international conference in pattern recognition},
  pages 351--353, 1982.

\bibitem{bmty05}
M.~F. Beg, M.~I. Miller, A.~Trouv\'e, and L~Younes.
\newblock Computing large deformation metric mappings via geodesic flows of
  diffeomorphisms.
\newblock {\em Int J. Comp. Vis.}, 61(2):139--157, 2005.

\bibitem{boo89}
L.~Bookstein, F.
\newblock Principal warps: Thin plate splines and the decomposition of
  deformations.
\newblock {\em IEEE trans. PAMI}, 11(6):567--585, 1989.

\bibitem{cy01}
V~Camion and L.~Younes.
\newblock Geodesic interpolating splines.
\newblock In M~Figueiredo, J~Zerubia, and K~Jain, A, editors, {\em EMMCVPR
  2001}, volume 2134 of {\em Lecture notes in computer sciences}. Springer,
  2001.

\bibitem{clz05}
M.~Chen, S.~Liu, and Y.~Zhang.
\newblock A two-component generalization of the camassa-holm equation and its
  solutions.
\newblock {\em Lett. Math. Phys.}, 75:1--15, 2005.

\bibitem{dgm98}
P~Dupuis, U~Grenander, and M~Miller.
\newblock Variational problems on flows of diffeomorphisms for image matching.
\newblock {\em Quaterly of Applied Math.}, 1998.

\bibitem{Falqui06}
G.~Falqui.
\newblock On a camassa-holm type equation with two dependent variables.
\newblock {\em J. Phys. A: Math. Gen.}, 39, 2006.

\bibitem{gy05}
L.~Garcin and L.~Younes.
\newblock Geodesic image matching: A wavelet based energy minimization scheme.
\newblock In {\em Proceedings of EMMCVPR 2005}, volume 3757 of {\em Lecture
  Notes in Computer Science}, pages 349--364, 2005.

\bibitem{gty04}
J~Glaun\`es, A~Trouv\'e, and L~Younes.
\newblock Diffeomorphic matching of distributions: A new approach for
  unlabelled point-sets and sub-manifolds matching.
\newblock In {\em Proceedings of CVPR'04}, 2004.

\bibitem{gvm03}
J~Glaun\`es, M~Vaillant, and I~Miller, M.
\newblock Landmark matching via large deformation diffeomorphisms on the
  sphere.
\newblock {\em Journal of Mathematical Imaging and Vision, MIA 2002 special
  issue}, (to appear) 2003.

\bibitem{gram99}
A~Guimon, A~Roche, N~Ayache, and J~Meunier.
\newblock Three-dimensional brain warping using the demons algorithm and
  adptive intensity corrections.
\newblock Technical report, INIRIA Sophia-Antipolis, 1999.

\bibitem{Ho2002}
D.~Holm, D.
\newblock Euler-poincar\'e dynamics of perfect complex fluids.
\newblock In P.~Newton, P.~Holmes, and A.~Weinstein, editors, {\em Geometry,
  Mechanics, and Dynamics: in honor of the 60th birthday of Jerrold E.
  Marsden}, pages 113--167. Springer, 2002.

\bibitem{HoMaRa1998}
D.~D. Holm, J.~E. Marsden, and T.~S. Ratiu.
\newblock The {Euler--Poincaré} equations and semidirect products with
  applications to continuum theories.
\newblock {\em Adv. in Math.}, 137:1--81, 1998.

\bibitem{jksj07}
S.~Joshi, A.~Klassen, E.and~Srivastava, and I.~Jermyn.
\newblock Removing shape-preserving transformations in square-root elastic
  (sre) framework for shape analysis of curves.
\newblock In Springer, editor, {\em Energy Minimization Methods in Computer
  Vision and Pattern Recognition, EMMCVPR 2007}, number 4679 in Lecture Notes
  in Computer Science, pages 387--398, 2007.

\bibitem{jm00}
S~Joshi and M~Miller.
\newblock Landmark matching via large deformation diffeomorphisms.
\newblock {\em IEEE transactions in image processing}, 9(8):1357--1370, 2000.

\bibitem{Ku2007}
A.~Kuz'min, P.
\newblock Two-component generalizations of the camassa-holm equation.
\newblock {\em Math. Notes}, 81:130--134, 2007.

\bibitem{mr99}
E~Marsden, J and S~Ratiu, T.
\newblock {\em Introduction to Mechanics and Symmetry}.
\newblock Springer, 1999.

\bibitem{mcm06}
R.~McLachlan and S.~Marsland.
\newblock Kelvin--helmholtz instability of momentum sheets in the euler
  equations for planar diffeomorphisms.
\newblock {\em SIAM J. Appl. Dyn. Sys.}, 5:726--758, 2006.

\bibitem{my01}
I~Miller, M and L~Younes.
\newblock Group action, diffeomorphism and matching: a general framework.
\newblock {\em Int. J. Comp. Vis}, 41:61--84, 2001.
\newblock ({\it Originally published in electronic form in: Proceeding of SCTV
  99, http://www.cis.ohio-state.edu/~szhu/SCTV99.html}).

\bibitem{msj05}
W.~Mio, A.~Srivastava, and S.~Joshi.
\newblock On the shape of plane elastic curves.
\newblock Technical report, Department of Mathematics, Florida State Univ.,
  2005.

\bibitem{qym07}
A.~Qiu, L.~Younes, and I.~Miller, M.
\newblock Intrinsic and extrinsic analysis in computational anatomy.
\newblock {\em Neuroimage}, 2007.
\newblock In press.

\bibitem{tog99}
W~Toga, A, editor.
\newblock {\em Brain warping}.
\newblock Academic Press, 1999.

\bibitem{tro98}
A.~Trouv\'e.
\newblock Diffeomorphism groups and pattern matching in image analysis.
\newblock {\em Int. J. of Comp. Vis.}, 28(3):213--221, 1998.

\bibitem{ty00}
A.~Trouv\'e and L.~Younes.
\newblock Diffeomorphic matching in 1d: designing and minimizing matching
  functionals.
\newblock In D.~Vernon, editor, {\em Proceedings of ECCV 2000}, 2000.

\bibitem{ty01}
A.~Trouv\'e and L.~Younes.
\newblock On a class of optimal matching problems in 1 dimension.
\newblock {\em Siam J. Control Opt.}, 39(4):1112--1135, 2001.

\bibitem{ty05}
A.~Trouv\'e and L~Younes.
\newblock Local geometry of deformable templates.
\newblock {\em SIAM J. Math. Anal.}, 37(1):17--59, 2005.

\bibitem{ty05b}
A.~Trouv\'e and L.~Younes.
\newblock Metamorphoses through lie group action.
\newblock {\em Found. Comp. Math.}, pages 173--198, 2005.

\bibitem{vg05}
M.~Vaillant and J.~Glaun\`es.
\newblock Surface matching via currents.
\newblock In Springer, editor, {\em Proceedings of Information Processing in
  Medical Imaging (IPMI 2005)}, number 3565 in Lecture Notes in Computer
  Science, 2005.

\bibitem{vmty04}
M~Vaillant, I~Miller, M, A~Trouv\'e, and L~Younes.
\newblock Statistics on diffeomorphisms via tangent space representations.
\newblock {\em Neuroimage}, 23(S1):S161--S169, 2004.

\bibitem{wbrc07}
L.~Wang, F.~Beg, M., T.~Ratnanather, J., C.~Ceritoglu, L.~Younes, J.~Morris,
  J.~Csernansky, and I.~Miller, M.
\newblock Large deformation diffeomorphism and momentum based hippocampal shape
  discrimination in dementia of the alzheimer type.
\newblock {\em IEEE Transactions on Medical Imaging}, 26(462-470), 2006.

\bibitem{you98}
L.~Younes.
\newblock Computable elastic distances between shapes.
\newblock {\em SIAM J. Appl. Math}, 58(2):565--586, 1998.

\bibitem{ymsm08}
L.~Younes, P.~Michor, J.~Shah, and D.~Mumford.
\newblock A metric on shape spaces with explicit geodesics.
\newblock {\em Rendiconti Lincei - Math. e Appl.}, 19(1):25--57, 2008.

\end{thebibliography}

\end{document}